\documentclass[journal]{IEEEtran}
\usepackage{caption}
\usepackage{float}
\usepackage{booktabs}
\usepackage{graphicx}
\usepackage{subfigure} 
\usepackage{amssymb}
\usepackage{bbding} 
\usepackage{indentfirst}
\usepackage{enumerate}
\usepackage{hyperref}
\usepackage{multirow}
\usepackage{makecell}  
\usepackage{nicematrix}
\usepackage[ruled]{algorithm2e}
\usepackage{colortbl}
\definecolor{mygray}{gray}{.95}

\makeatletter
\newcommand{\removelatexerror}{\let\@latex@error\@gobble}
\makeatother
\makeatletter
\renewcommand{\maketag@@@}[1]{\hbox{\m@th\normalsize\normalfont#1}}%
\makeatother

\hypersetup{
  colorlinks=true,
  citecolor=blue,
  linkcolor=red,
  urlcolor=blue
  }

\usepackage{tikz,xcolor}
\definecolor{lime}{HTML}{A6CE39}
\DeclareRobustCommand{\orcidicon}{%
	\begin{tikzpicture}
	\draw[lime, fill=lime] (0,0) 
	circle [radius=0.134] 
	node[white] {{\fontfamily{qag}\selectfont \tiny ID}};    \draw[white, fill=white] (-0.0625,0.095) 
	circle [radius=0.007];    \end{tikzpicture}
	\hspace{-2mm}}
\foreach \x in {A, ..., Z}{%
	\expandafter\xdef\csname orcid\x\endcsname{\noexpand\href{https://orcid.org/\csname orcidauthor\x\endcsname}{\noexpand\orcidicon}}
}

\begin{document}
	
\captionsetup[figure]{name={Fig.},labelsep=period}
\captionsetup[table]{name={TABLE},labelsep=period}

\title{Frequency-Assisted Mamba for Remote Sensing Image Super-Resolution}

\author{Yi~Xiao\orcidA{},~\IEEEmembership{Graduate Student Member,~IEEE,}
       Qiangqiang~Yuan\orcidB{}, ~\IEEEmembership{Member,~IEEE,}
       Kui~Jiang\orcidC{}, ~\IEEEmembership{Member,~IEEE,}
       Yuzeng~Chen\orcidG{}, 
       Qiang~Zhang\orcidD{},~\IEEEmembership{Member,~IEEE,}
       and~Chia-Wen~Lin\orcidE{}, ~\IEEEmembership{Fellow,~IEEE.} 
\thanks{This work was supported in part by the National Natural
	Science Foundation of China under Grant 423B2104 and in part by the Fundamental Research Funds for the Central Universities under Grant 2042024kf0020 and 2042023kfyq04. (\emph{Corresponding author: Qiangqiang Yuan; Kui Jiang}.)
 }
\thanks{Yi Xiao, Qiangqiang Yuan, and Yuzeng Chen are with the School of Geodesy and Geomatics, Wuhan University, Wuhan 430079, China (e-mail: xiao\_yi@whu.edu.cn; yqiang86@gmail.com; yuzeng\_chen@whu.edu.cn).}
\thanks{Kui Jiang is with the School of Computer Science and Technology, Harbin Institute of Technology, Harbin 150001, China (e-mail: jiangkui@hit.edu.cn).}
\thanks{Qiang Zhang is with the Information Science and Technology College, Dalian Maritime University, Dalian 116026, China (e-mail: qzhang95@dlmu.edu.cn).}
\thanks{Chia-Wen Lin is with the Department of Electrical Engineering and the Institute of Communications Engineering, National Tsing Hua University, Hsinchu 300, Taiwan (e-mail: cwlin@ee.nthu.edu.tw).}
}

%
%

\markboth{Accepted to IEEE Transactions on Multimedia}%
{Shell \MakeLowercase{\textit{et al.}}: Bare Demo of IEEEtran.cls for IEEE Journals}
%



\maketitle
\begin{abstract}
Recent progress in remote sensing image (RSI) super-resolution (SR) has exhibited remarkable performance using deep neural networks, \emph{e.g.,} Convolutional Neural Networks and Transformers. However, existing SR methods often suffer from either a limited receptive field or quadratic computational overhead, resulting in sub-optimal global representation and unacceptable computational costs in large-scale RSI. To alleviate these issues, we develop the first attempt to integrate the Vision State Space Model (Mamba) for RSI-SR, which specializes in processing large-scale RSI by capturing long-range dependency with linear complexity. To achieve better SR reconstruction, building upon Mamba, we devise a Frequency-assisted Mamba framework, dubbed FMSR, to explore the spatial and frequent correlations. In particular, our FMSR features a multi-level fusion architecture equipped with the Frequency Selection Module (FSM), Vision State Space Module (VSSM), and Hybrid Gate Module (HGM) to grasp their merits for effective spatial-frequency fusion. Considering that global and local dependencies are complementary and both beneficial for SR, we further recalibrate these multi-level features for accurate feature fusion via learnable scaling adaptors. Extensive experiments on AID, DOTA, and DIOR benchmarks demonstrate that our FMSR outperforms state-of-the-art Transformer-based methods HAT-L in terms of PSNR by 0.11 dB on average, while consuming only 28.05\% and 19.08\% of its memory consumption and complexity, respectively. Code will be available at \url{https://github.com/XY-boy/FreMamba}
\end{abstract}


%
\IEEEpeerreviewmaketitle

\section{Introduction}

\IEEEPARstart{H}{igh}-resolution remote sensing imagery (RSI), which records high-quality earth observation details, provides promising prospects for large-scale and fine-grained applications \cite{tmm1, chen2024changemamba, chen2023exchange, wjj2024isprs, tmm3, guo2024saan, du2023global}. However, the complex imaging environment (\emph{e.g.,} scattering and tremor) often impedes high-resolution (HR) image acquisition \cite{liu2023rethinking, zhang2023hyperspectral, chen2021hybrid, zhang2022cooperated}. Moreover, RSI often undergoes severe compression and downsampling to tame the transmission instability between satellites and ground stations, resulting in suboptimal scene representation. Hence, reconstructing HR images from low-resolution (LR) observations is crucial for improving both human perception and subsequent applications.

\begin{figure}[!t]
\centering
\includegraphics[width=3.5in]{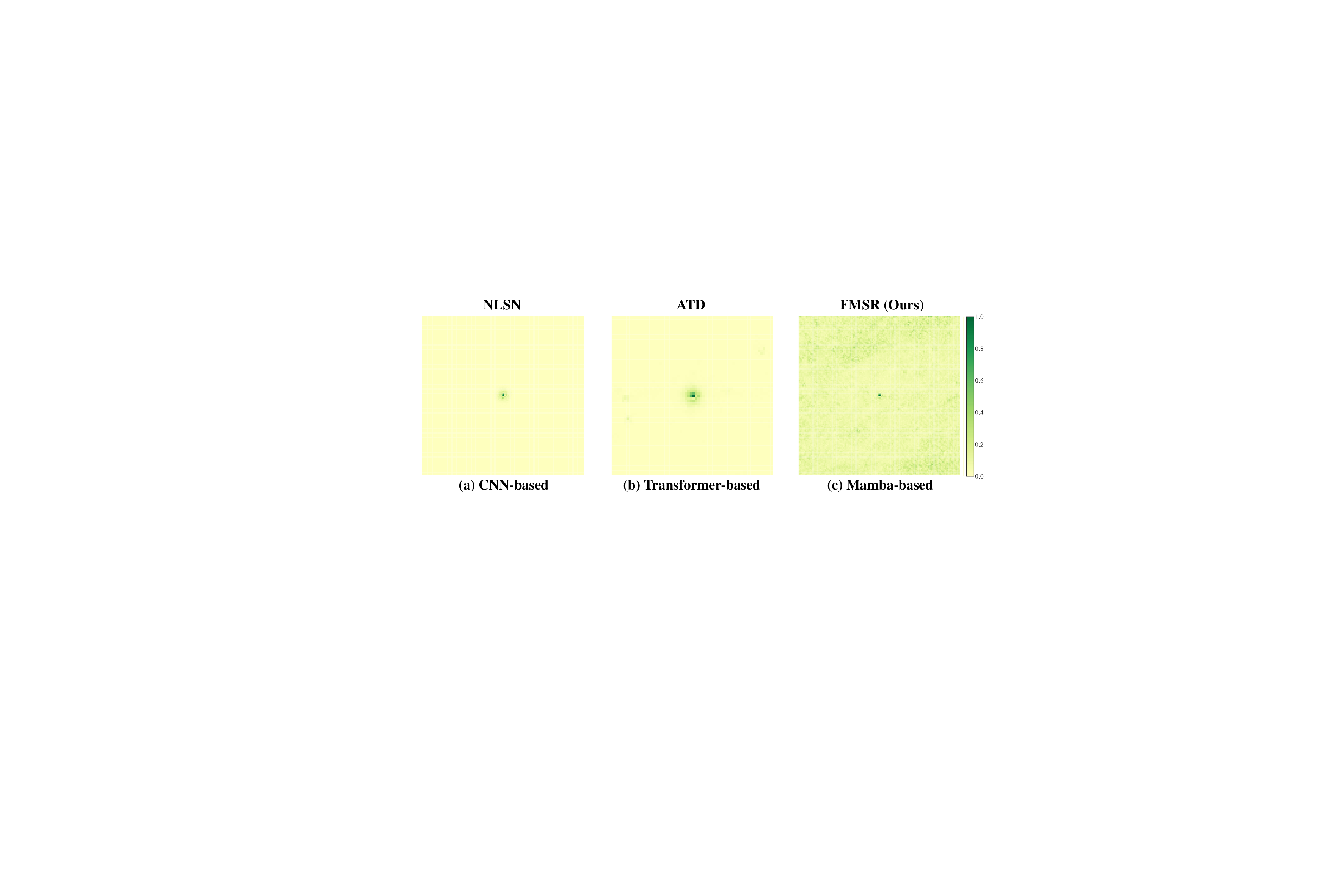}%
\captionsetup{font={scriptsize}}   
\caption{The Effective Receptive Field (ERF) \cite{erf} comparison for (a) CNN-based method NLSN \cite{nlsa}, (b) Transformer-based model ATD \cite{ATD}, and the proposed Mamba-based network FMSR. A wider distribution of dark areas demonstrates larger ERF. Our FMSR effectively obtains the largest ERF, indicating favorable global exploration capability.}
\label{introduction}
\end{figure}

In contrast to upgrading hardware maintenance, super-resolution (SR) techniques provide a flexible and cost-effective alternative by predicting latent HR images from their LR counterparts \cite{tmm5, tmm6, xiao2020space, xiao2021space}. Early efforts often relied on hand-crafted priors to tame the ill-posedness \cite{kim2010single, tmm7, zhou2024PDR, zhou2023adp, xiao2023cutmib}. However, they struggled to produce accurate results and involved laborious optimization processes. Recently, deep neural networks have demonstrated remarkable progress in SR tasks and achieved superior performance over traditional approaches, such as Convolutional Neural Networks (CNNs) and Transformers. While CNN-based methods commonly invent elaborate attention mechanisms to grasp the informative features, restrained by the inherent nature of convolution units, they have limited respective fields and cannot capture long-range dependencies. As shown in Fig. \ref{introduction}, the effective receptive field (ERF) \cite{erf} of CNN-based model NLSN \cite{nlsa} is limited. This requirement is essential for SR tasks, as a predicted pixel needs prior knowledge from its surrounding region to be super-resolved, especially in wide-range RSI.

Transformer-based methods achieve increased respective fields by leveraging global interaction among all input data through a self-attention (SA) mechanism, demonstrating impressive performance across various domains \cite{han2022survey, Zheng_2024_CVPR, su2024high, tmm10, Chen_2023_CVPR, hou2024linearly}. Despite achieving superior performance against CNN-based approaches, these methods exhibit quadratic complexity with respect to the token size. In this context, taming Transformer for high-resolution scenarios presents a significant challenge, particularly for large-scale remote sensing images. Although some approaches seek a lightweight SA for global modeling, such as recursive SA \cite{rgt} and window-based SA \cite{hat}, they usually come at the expense of global modeling accuracy and require stacking many blocks to establish a global dependency, thus increasing the computational budget. Moreover, the inherent issue of quadratic complexity remains unsolved. Therefore, a natural question arises: \emph{can a more efficient yet effective solution be developed to grasp the long-range dependencies across large-scale RSI?}

The recent-popular State Space Model (SSM) could be a promising answer to this question. Originating from Kalman filtering \cite{kalman1960new}, SSM primarily employs linear filtering and prediction methods to represent the evolution of the internal state of the system, thus naturally enjoying linear complexity. By integrating SSM with the MultiLayer Perceptron (MLP) block of Transformer, the simplified architecture Mamba \cite{guo2022visual} is achieved, which introduces a selection scanning mechanism in SSM to filter out irrelevant information for long-sequence modeling. Recently, Mamba demonstrated impressive results in various domains, making it a possible replacement for the Transformer model. Nevertheless, although Mamba exhibits favorable performance and can serve as an alternative to Transformer, some potential problems persist in large-scale earth observation scenarios, making introducing Mamba into RSI SR more challenging. \textbf{Firstly, images captured by satellite platforms often lose crucial frequency information for perception, which requires heterogeneous representation for accurate reconstruction.} The underlying reason lies in that the original Mamba processes each token equally in the spatial domain, limiting its ability to perceive informative frequency signals across entire images. \textbf{Secondly, there exists spatial diversity among observed objects in RSI, which is barely explored.} While Mamba is effective in exploring long-range dependencies, it lacks explicit consideration of spatially-varying contents, resulting in suboptimal pixel-wise representation during local modeling.

To mitigate the aforementioned problem, we first attempt to extend Mamba from the perspective of frequency analysis and propose a frequency-assisted Mamba for RSI SR, termed FMSR. Specifically, instead of solely employing VSSM for long-range modeling, we devise an effective Frequency Selection Module (FSM) to adaptively identify informative frequency cues vital for perception. By incorporating FSM with Mamba, the resulting frequency-assisted mamba block can better utilize the complementary strengths between Mamba and frequency analysis for accurate SR. Considering that the VSSM aggregates image features via patch-wise linear scaling, it inevitably overlooks some pixel-wise localities. To address this, we further develop a Hybrid Gate Module (HGM) to better introduce a local inductive bias. Unlike the commonly used channel attention, HGM allows for selective amplification or attenuation of local features on spatial position, effectively enhancing spatially-varying representation during channel-wise correction learning. 

Furthermore, there is inherent misalignment between different level features (global and local). The direct fusion of multi-level features inevitably arises in confused and conflicted representation. To this end, we introduce a learnable adapter to rescale cross-level representation for improved integration. Overall, equipped with the above designs, our FMSR can capture both global and local dual-domain dependencies for RSI SR while maintaining moderate complexity.

Our main contribution can be summarized as follows:
\begin{enumerate}[1)]
\item We introduce the first state space model for remote sensing image super-resolution (FMSR), highlighting Mamba's capability for efficient and effective global modeling in large-scale remote sensing scenarios.
\item To integrate more high-frequency cues into Mamba, we develop a Frequency Selection Module (FSM), which adaptively identifies and selects the most informative frequency signals during the fast fourier transformation process.
\item We design a Hybrid Gate Module (HGM) that integrates the local bias of CNN operators with spatially-varying coordinates to enhance the locality of feature representation, leading to more accurate and faithful SR performance.
\end{enumerate}

\par The remainder of this paper is organized as follows. Section \ref{Section2} reviews the related knowledge pertinent to our FMSR. In Section \ref{section3}, we provide detailed descriptions of the implementation of our FMSR. Section \ref{section4} contains extensive experiments conducted on widely used remote sensing benchmarks. Section \ref{section5} concludes our work.

\section{Related Work} \label{Section2}
In this section, we first present a comprehensive review of remote sensing image super-resolution. Then, we introduce relevant background knowledge for this study, including state space models and frequency learning.

\subsection{Remote Sensing Image Super-resolution}
RSI SR has witnessed significant advancements with the booming of deep learning \cite{su2022global, huang2023pami, FCCL_CVPR22, ediffsr, chen2023cross}. The primary focus of this task lies in extracting prior knowledge from LR images, which can be broadly categorized into three categories: CNN-, Transformer-, and Mamba-based methods.

\textbf{CNN-based}. Drawing inspiration from SRCNN \cite{srcnn}, early efforts usually elaborate CNNs with advanced modules, such as residual \cite{vdsr} and dense structure \cite{rdn} and attention mechanisms \cite{rcan, han}. Mei et al. \cite{nlsa} introduced non-local sparse attention to capture global dependencies inherent in LR images. Similarly, Lei et al. \cite{hsenet} extended the non-local mechanism by exploring cross-scale similarity in RSI. While these methods improve the local receptive field nature of CNNs, they suffer from significant computational overhead during non-local exploration, making them less efficient in large-scale remote sensing scenes. Moreover, limited by the local bias of CNN, they cannot capture critical long-range dependencies and reach a plateau in performance.

\textbf{Transformer-based}. The core insight of Transformer lies in the Self-Attention (SA) mechanism \cite{chen2023msdformer}, which has demonstrated superior long-range modeling capability and outperformed CNN-based methods. Lei et al. \cite{transenet} devised a multi-stage Transformer-enhanced network. Recently, Chen et al. \cite{hat} proposed to activate more pixels in SR tasks by combining both CNN and self-attention. More Recently, an improved network \cite{grl} was proposed by integrating channel attention, improved SA, and anchored stripe attention. However, due to the quadratic complexity of SA regarding token size, they are less efficient in handling high-resolution images, large-scale RSI in particular. To alleviate the computational budget of SA, Chen et al. \cite{rgt} proposed a recursive SA by recursively aggregating input features for enriched token representation. Nevertheless, efficient SA often sacrifices global modeling capability, and despite efforts to mitigate the quadratic complexity, the inherent problem of SA remains unsolved.

\textbf{Mamba-based.} In light of the success of Mamba, some scholars \cite{mambair} attempt to introduce Mamba for efficient global modeling with linear complexity. However, to the best of our knowledge, the potential of Mamba in RSI SR remains unexplored. Since the imaging is wide-ranging, the content in RSI exhibits complex and diverse properties. Furthermore, compared to natural images, the texture information of RSI is less prominent, and the vital high-frequency information tends to vanish in deep models.

In summary, there is an urgent need for an efficient yet effective scheme to model the heterogeneous representation and to seek a practical solution to explore the critical high-frequency components. This paper pioneers exploring Mamba's potential in the RSI SR task and extends Mamba with frequency analysis, providing an effective and efficient paradigm for this challenging issue.

\begin{figure*}[!t]
\centering
\includegraphics[width=7in]{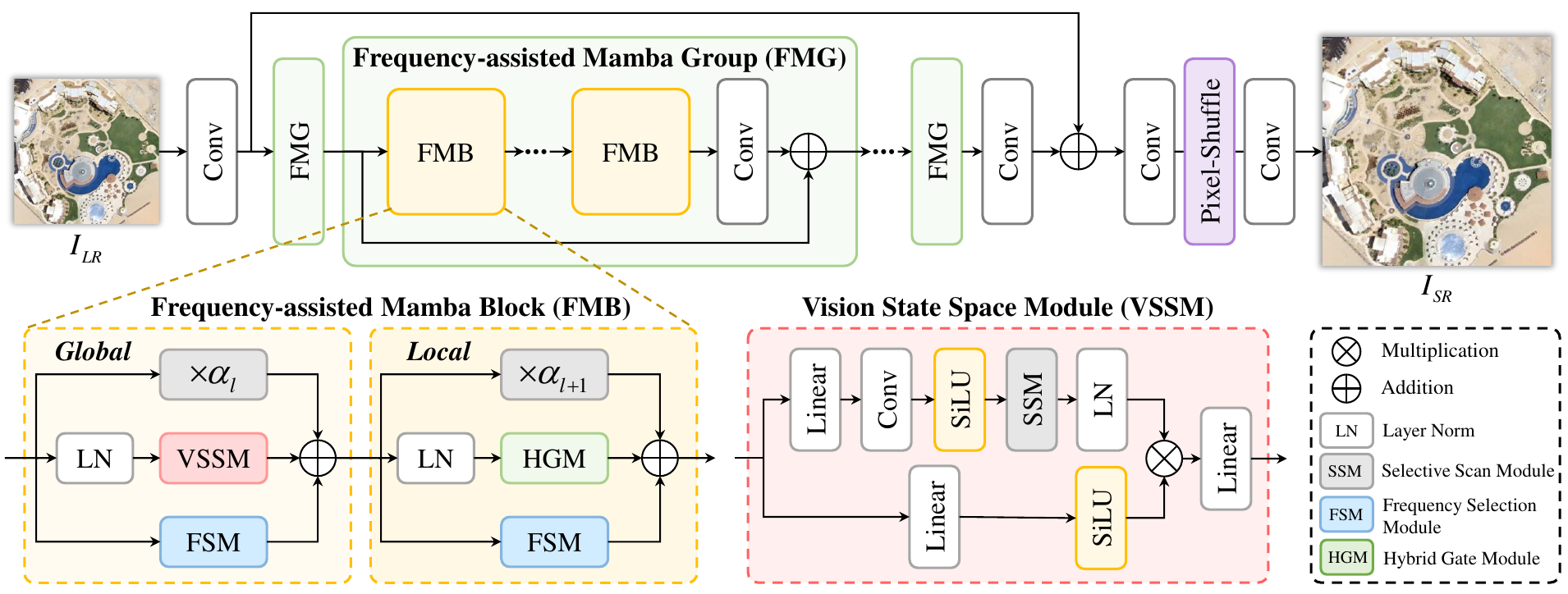}%
\captionsetup{font={scriptsize}}   
\caption{Overview of the proposed FMSR. The Frequency-assisted Mamba Blocks (FMB) are arranged sequentially in Frequency-assisted Mamba Groups (FMG). In FMB, a Frequency Selection Module (FSM) is adopted to assist the learning process of the Vision State Space Module (VSSM) and Hybrid Gate Module (HGM). $\alpha_l$ is a learnable adaptor for hybrid adaptive integration in the $l$-th FMB. }
\label{network}
\end{figure*}

\subsection{State Space Model}
Recently, state space models (SSMs) \cite{smith2022simplified} have emerged as a promising approach, demonstrating competitive performance in long-range modeling compared to transformers. The key advantage of SSMs lies in their linear scaling with sequence length, providing a global perspective with linear complexity. Gu et al. \cite{gu2022efficiently} pioneered the SSM to tackle long-sequence data modeling, illustrating promising linear scaling properties. Subsequently, they put forward a variant named Mamba \cite{mamba}, which adopts a selective mechanism and efficient network design. Mamba has shown superior performance to transformers in natural language processing tasks. In light of the success of Mamba, it has been introduced in computer vision tasks and demonstrated impressive performance, including object detection \cite{dong2024fusionmamba}, image classification \cite{liu2024vmamba}, and biomedical image segmentation \cite{ma2024umamba}. However, research on Mamba in low-level vision tasks is still in its primary stage, and efforts for RSI SR remain unexplored.

This paper adapts Mamba for the SR task. Unlike previous works that solely replace self-attention with the Vision State Space Model (VSSM) for long-range modeling, we promote the perspective of global exploration in the spatial-frequency dual domain. Compared to spatial-wise modeling, our method is more effective by incorporating latent high-frequency cues for better global representation.

\begin{figure}[!t]
\centering
\includegraphics[width=3.4in]{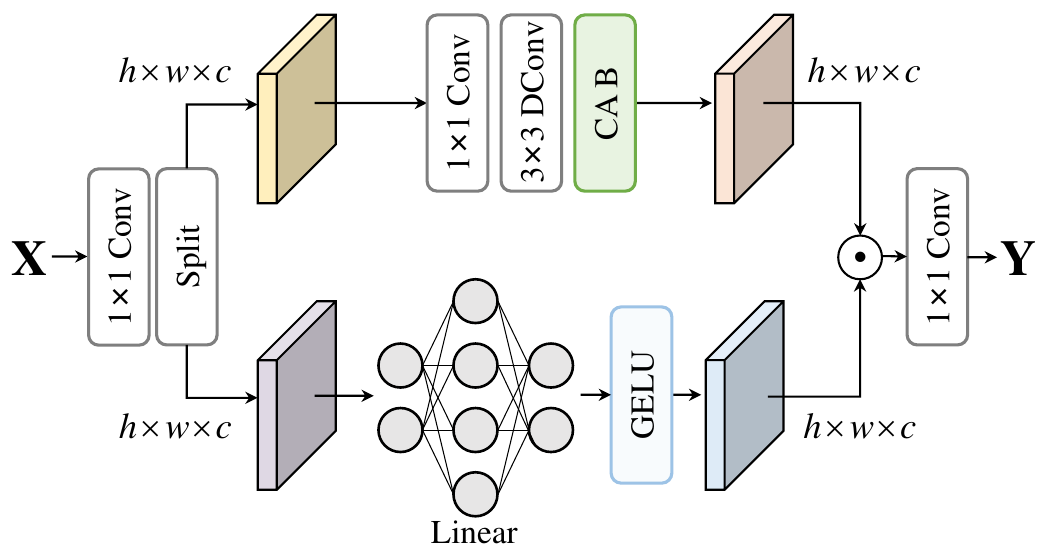}%
\captionsetup{font={scriptsize}}   
\caption{The proposed Hybrid Gate Module (HGM) conceptual illustration. The input feature \textbf{X} is split in the channel dimension and fed through a Channel Attention Block (CAB) and a pixel-wise linear projection layer, respectively. After a Hadamard product operation, a $1\times1$ convolution generates the output tensor \textbf{Y}.}
\label{hgm}
\end{figure}

\begin{figure}[!t]
\centering
\includegraphics[width=3.3in]{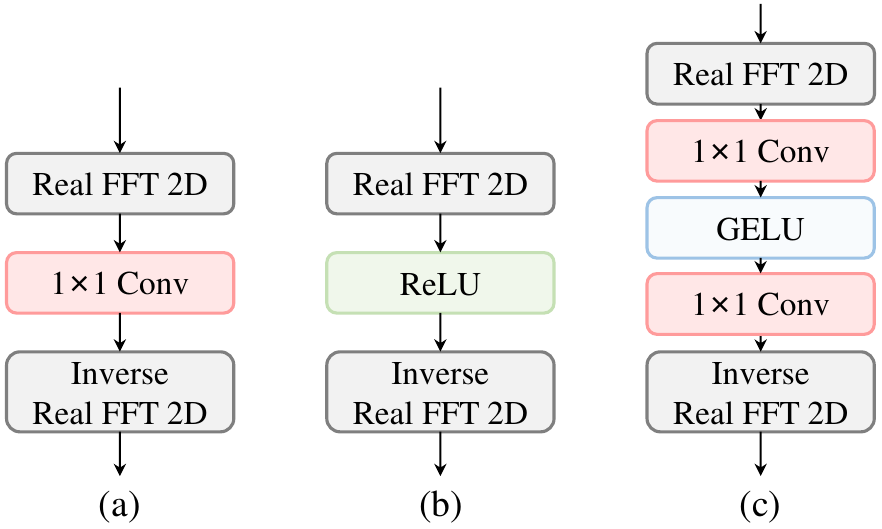}%
\captionsetup{font={scriptsize}}   
\caption{Three variants of Frequency Selection Module (FSM). Here, we adopt 2D Fast Fourier Transformation (FFT) for frequency learning.} 
\label{fsm}
\end{figure}

\subsection{Fourier Transform}
The Fast Fourier transform (FFT) can be viewed as a global statistical signal, making it suitable for global information analysis. In light of this, various visual tasks leverage FFT for frequency domain modeling, such as semantic segmentation \cite{yang2020fda} and image classification \cite{xu2021fourier}. In low-level vision scenes, Mao et al. \cite{mao2021deep} introduced a residual FFT module capable of capturing comprehensive local high-frequency details for low-light enhancement. Li et al. \cite{li2023embedding} integrated the Fourier transform into a deep network to mitigate noise amplification during luminance enhancement. Guo et al. \cite{guo2023spatial} proposed a window-based frequency channel attention mechanism. Wang et al. \cite{wang2023spatial} conducted mutual learning of frequency and spatial domains to improve face image SR. Further, some approaches incorporate FFT into the loss function to enhance reconstructed sharp details \cite{cho2021rethinking}. 

However, these methods often prioritize elaborate networks for exploring frequency signals, neglecting frequency contribution analysis. This inevitably amplifies harmful frequencies and increases computational overhead. In this study, we dynamically adjust input frequency-domain features using lightweight activation weights and a convolutional layer to emphasize informative frequencies for accurate SR, which offers more flexibility to modulate selection thresholds.

\section{Methodology}\label{section3}
In this section, we introduce the implementation details of our FMSR. The FMSR comprises a straightforward backbone with convolution layers, Frequency-assisted Mamba Groups (FMG), and a pixel-shuffle layer. We start with an overview of FMSR, and then we dive into its components by explaining: the Frequency-assisted Mamba Block (FMB), Vision State Space Module (VSSM), Hybrid Gate Module (HGM), and Frequency Selection Module (FSM).

\subsection{Overview of FMSR}
As illustrated in Fig. \ref{network}, the proposed FMSR consists of three major stages: shallow feature extraction of LR, deep feature acquisition, and reconstruction of HR. Firstly, the given LR input $I_{LR}$ is fed into a $3\times3$ convolution layer $\phi$ with learnable parameters $\theta$ to generate initial feature $F_0$, which can be written as:
\begin{equation}
{F_0} = \phi ({I_{LR}},\theta ).
\end{equation}
Subsequently, the $F_0$ undergoes deep feature extraction with multiple Frequency-assisted Mamba Groups (FMGs), followed by a $3\times3$ convolution for feature refinement. This process can be expressed as:
\begin{equation}
{F_m} = {\rm FMG}_m({\rm FMG}_{m - 1}( \cdots {\rm FMG}_1({F_0}))),
\end{equation}
where $m$ represents the number of FMG and $F_m$ is the output of $m$-th FMG. We incorporate a global skip connection to prepare high-quality features $F_{rec}$ for reconstruction, \emph{i.e.,} ${F_{rec}} = {\rm Conv}({F_m}) + {F_0}$. Finally, a $3\times3$ convolution, a pixel-shuffle layer $\rm PS$, and a terminal convolution are employed to upscale and restore the super-resolved output $I_{SR}$:
\begin{equation}
{I_{SR}} = {\rm Conv}({\rm PS}({\rm Conv}({F_{rec}}),s)),
\end{equation}
where $s$ means the upscale factor.

\subsection{Frequency-assisted Mamba Block}
The structure of FMB is shown in Fig. \ref{network}, from which we can find that FMB serves as a primary component in FMG. In particular, each FMG contains cascaded FMB, a convolution layer, and a residual connection. The FMB is responsible for exploring global and local representations in a frequency-spatial dual domain. The function of $i$-th FMG can be summarized as follows:
\begin{equation}
{F_i} = {\rm Conv}(\Psi_n(\Psi_{n - 1}( \cdots \Psi({F_{i - 1}})))) + {F_{i - 1}},
\end{equation}
where $\Psi_n$ denotes the $n$-th FMB in each FMG.

Specifically, FMB performs \emph{global} and \emph{local} modeling in the frequency-spatial dual domain. Given output of $\Psi_{l-1}$, termed $x_{l-1}$, the \emph{l}-th FMB $\Psi_{l}$ processes $x_{l-1}$ with three parallel branches to grasp their merits of global representations: 1) A Layer Norm (LN) followed by the 2D Vision State Space Module (VSSM) is developed to capture the spatial-wise long-term information, 2) a Frequency Selection Module (FSM) is equipped to introduce more high-frequency cues in the frequency domain while promoting the capability of VSSM, 3) a learnable scaling factor $\alpha_l$ for dynamic feature aggregation.
The feature $y$ after global frequency-spatial exploration can be formulated as:
\begin{equation}
y = {\alpha _l} \cdot {x_{l - 1}} + {\rm VSSM} ({\rm LN} ({x_{l - 1}})) + {\rm FSM} ({x_{l - 1}}).
\end{equation}
After that, $y$ undergoes further local modeling. Similarly, we adopt LN to normalize $y$ and then use a Hybrid Gate Module (HGM) to grasp the spatial locality. Also, FSM is employed to assist the local modeling process with frequency learning. Finally, another scale factor $\alpha_{l+1}$ is used to adaptively integrate the output from local and global representations, which can be expressed as:
\begin{equation}
x_l = {\alpha _{l+1}} \cdot y + {\rm HGM} ({\rm LN} (y)) + {\rm FSM} (y).
\end{equation}

\begin{table*}[!t]
  \centering
  \captionsetup{font={scriptsize}, labelsep = newline, justification=centering}
  \caption{Ablation Studies of different components on the proposed FMSR. The best and second-best PSNR performances are highlighted in \textbf{bold} and \underline{underlined}. Note that we use residual blocks in Model-1 and Model-2 as feed-forward networks.}
  \setlength{\tabcolsep}{4mm}{
    \begin{tabular}{ccccccc}
    \toprule[1.5pt]
    Components & \multicolumn{1}{c}{Model-1 (Base)} & \multicolumn{1}{c}{Model-2} & \multicolumn{1}{c}{Model-3} & \multicolumn{1}{c}{Model-4} & \multicolumn{1}{c}{Model-5} & \multicolumn{1}{c}{Model-6 (FMSR)} \\
    \midrule
    \midrule
    Window-based Self-Attention \cite{hat}   & \Checkmark     & \XSolidBrush      &  \XSolidBrush     & \XSolidBrush      & \XSolidBrush      &\XSolidBrush  \\
    2D-VSSM (Ours)  &  \XSolidBrush     & \Checkmark     & \Checkmark     & \Checkmark     & \Checkmark     & \Checkmark \\
    MLP Layer \cite{rgt}   &  \XSolidBrush     &  \XSolidBrush     & \Checkmark     &   \XSolidBrush    &  \XSolidBrush     & \XSolidBrush \\
    Channel Attention \cite{rcan}    &  \XSolidBrush     & \XSolidBrush      & \XSolidBrush      & \Checkmark     & \XSolidBrush      & \XSolidBrush  \\
    Hybrid Gate Module (Ours)   &  \XSolidBrush     &  \XSolidBrush     &  \XSolidBrush     &  \XSolidBrush     & \Checkmark     & \Checkmark \\
    Frequency Selection Module (Ours)   &  \XSolidBrush     & \XSolidBrush      &  \XSolidBrush     & \XSolidBrush      & \XSolidBrush      & \Checkmark \\
    \midrule    
    \midrule
    PSNR (dB) &  27.751     & 27.846      & 28.104      & 28.088      & \underline{28.122}      & \textbf{28.178} \\
    \bottomrule[1.5pt]
    \end{tabular}%
    }
\label{tb1}%
\end{table*}%

\subsection{Vision State Space Module}
Previous efforts often rely on Transformers to explore global dependency, which calculate the long-range response with the self-attention mechanism. Despite achieving favorable performance, they suffer from high complexity, hindering the efficient modeling in large-scale remote sensing images. Inspired by the success of the vision state space module in long-term modeling and aggregation with linear complexity, we first introduce VSSM to the RSI SR task.

In particular, as illustrated in Fig. \ref{network}, the normalized feature $x_N={\rm LN}(x_l)$ is expanded along the channel dimension by a linear projection operation $\phi_1$ with an expansion factor $\lambda$. Then, a series of operations including a $1\times1$ Depth-Wise Convolution (DWConv) $f_{1\times1}$, a SiLU activation $\sigma_1$, as well as the 2D-selective scan module (SSM) and LN are sequentially stacked to generate the output of the first branch, denoted $h_1$. This branch can be defined as: 
\begin{equation}
{h_1} = {\rm LN}({\rm SSM}(\sigma_1 ({f_{1 \times 1}}(\phi_1 ({x_N, \lambda}))))).
\end{equation}
In the second branch, another linear layer $\phi _2$ and SiLU function $\sigma _2$ are used. The output of this branch can be obtained by:
\begin{equation}
{h_2} = {\sigma _2}({\phi _2}({x_N},\lambda )). 
\end{equation}
Finally, to produce the final output $h_{out}$, the output $h_1$ and $h_2$ are incorporated via Hadamard product, followed by a linear layer $\phi _3$. That is:
\begin{equation}
{h_{out}} = {\phi _3}({h_1} \otimes {h_2}).
\end{equation}

\begin{figure}[!t]
\centering
\includegraphics[width=3in]{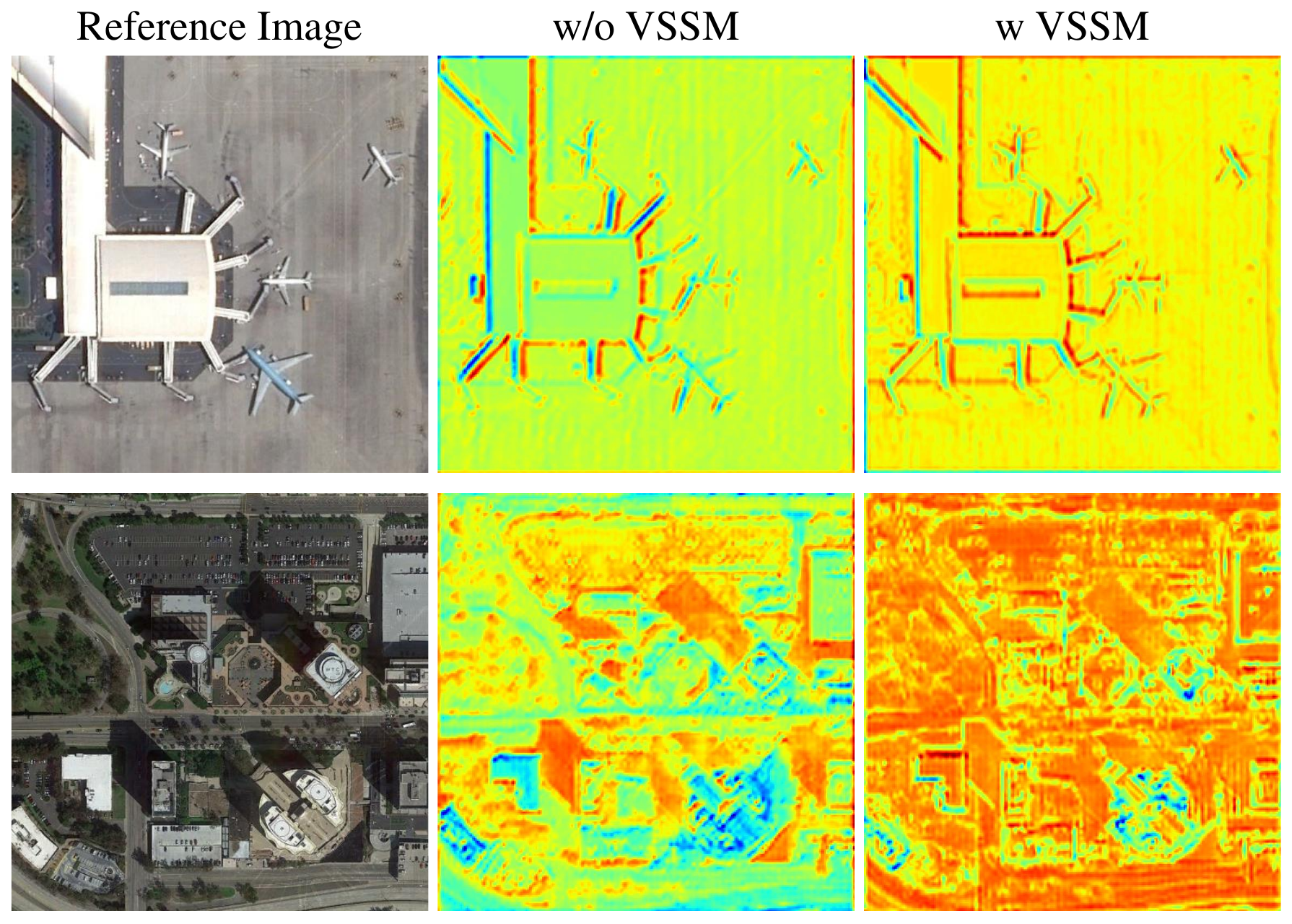}%
\captionsetup{font={scriptsize}}   
\caption{Feature visualization comparisons. The feature maps corresponding to each reference image are the results of the 56-\emph{th} channels in the final FMG. }
\label{fea-vis}
\end{figure}

\begin{table}[!t]
  \centering
  \captionsetup{font={scriptsize}, labelsep = newline, justification=centering}
  \caption{Ablation studies on different variants of Frequency Selection Module (FSM) as illustrated in Fig. \ref{fsm}. The best and second-best PSNR performances are highlighted in \textbf{bold} and \underline{underline}, respectively.}
\setlength{\tabcolsep}{3.5mm}{
    \begin{tabular}{cccc}
\toprule[1.5pt]
    Variants & FSM (a) & FSM (a)   & FSM (a)   \\
\midrule    
    \midrule
    PSNR (dB)& 	28.129      & \underline{28.147}     & \textbf{28.178}   \\
\bottomrule[1.5pt]
    \end{tabular}%
}
  \label{abla-fsm}%
\end{table}%

\begin{table}[!t]
  \centering
  \captionsetup{font={scriptsize}, labelsep = newline, justification=centering}
  \caption{Ablation studies on Hybrid Adaptive Integration (HAI) of global and local representations. The best and second best PSNR/SSIM performance are highlighted in \textbf{bold} and \underline{underline}, respectively. We report these results on AID \cite{aid}.}
    \setlength{\tabcolsep}{2mm}{
    \begin{tabular}{cccccc}
    \toprule[1.5pt]
    Method &  Skip & Adaptive $\alpha$ & \#Param. & PSNR (dB)  & SSIM \\
    \midrule    
    \midrule
    w/o HAI &       &       &   11.75M & \underline{30.80}    & \underline{0.8121} \\
    w/ Skip &   \Checkmark    &       &  11.75M& 30.57     & 0.7986 \\
    w/ HAI &    \Checkmark   &   \Checkmark    &  11.76M& \textbf{30.93}     & \textbf{0.8156} \\
    \bottomrule[1.5pt]
    \end{tabular}%
    }
  \label{abla-hai}%
\end{table}%

\subsection{Hybrid Gate Module}
Current methods often utilize either MLP layers \cite{hat} for feature propagation or incorporate convolution operations, such as attention mechanisms \cite{mambair}, after long-range exploration to introduce critical locality for improved performance. Our approach draws inspiration from these methods but condenses them into a unified hybrid module. To learn a comprehensive local context, the Hybrid Gate Module (HGM) incorporates the spatially-varying properties of remote sensing images (RSI) by selectively amplifying or attenuating local features in the pixel domain while preserving channel-specific features.

As shown in Fig. \ref{hgm}, HGM treats features captured by local convolution as coordinates and then multiplies them with a pixel-wise gating mask of the same size. Specifically, the input feature $\rm X$ is first processed through a $1\times1$ convolution to expand the channel dimension to $2c$. We then split $\rm X$ into two parts, $\rm X_1$ and $\rm X_2$, by halving the channel dimension. $\rm X_1$ and $\rm X_2$ are subsequently sent into the first and second branches, respectively. In the first branch, a $1\times1$ convolution, a $3\times3$ depth-wise convolution, and a channel attention block \cite{rcan} are used to yield the coordinates. 
\begin{equation}
{\rm X}_{coor} = {\rm CA}({\rm Dconv}({\rm Conv}({\rm X}_1))).
\end{equation}
In the second branch, we fed the feature through pixel-wise linear projection, somewhat similar to the MLP layer. In contrast to the MLP layer, we only activate the feature at the end of linear projection with the GELU activation function to generate the gate weights:
\begin{equation}
{\rm M} = {\rm GELU}({\rm Linear}({\rm X}_2)).
\end{equation}
Finally, the output $\rm Y$ can be obtained by:
\begin{equation}
{\rm Y} = {\rm Conv}({\rm M} \odot {{\rm X}_{coor}}).
\end{equation}

\subsection{Frequency Selection Module}
To achieve frequency-spatial dual domain representation at global and local levels, we equip VSSM and HGM with frequency exploration. As illustrated in Fig. \ref{fsm}, we devise three variants of frequency selection operations using Fast Fourier transformation (FFT):

(a) Using 2D real FFT and using a $1\times1$ convolution layer before inverse FFT, which means we do not perform frequency selection: 
\begin{equation}
{\rm Z} = {{\cal F}^{ - 1}}({\rm Conv}({\rm{{\cal F}}}(x))).
\end{equation}

(b) Inserting ReLU activation between FFT and Inverse FFT to dynamically select the frequency pattern:
\begin{equation}
{\rm Z} = {{\cal F}^{ - 1}}({\rm ReLU}({\rm{{\cal F}}}(x))).
\end{equation}

(c) Applying two stacks of $1\times1$ convolution layer and GELU activation function:
\begin{equation}
{\rm Z} = {{{\cal F}}^{ - 1}}{\rm Conv}(({\rm GELU}({\rm Conv}({\cal F}(x))))).
\end{equation}

We finally choose scheme (c) as our FSM as $1\times1$ convolution lets the network modulate ﬂexible thresholds for frequency selection with lightweight design. Ablation experiments demonstrate the effectiveness of (c) compared to (a) frequency analysis without selection and (b) ReLU-based selection.

\begin{figure}[!t]
\centering
\includegraphics[width=3in]{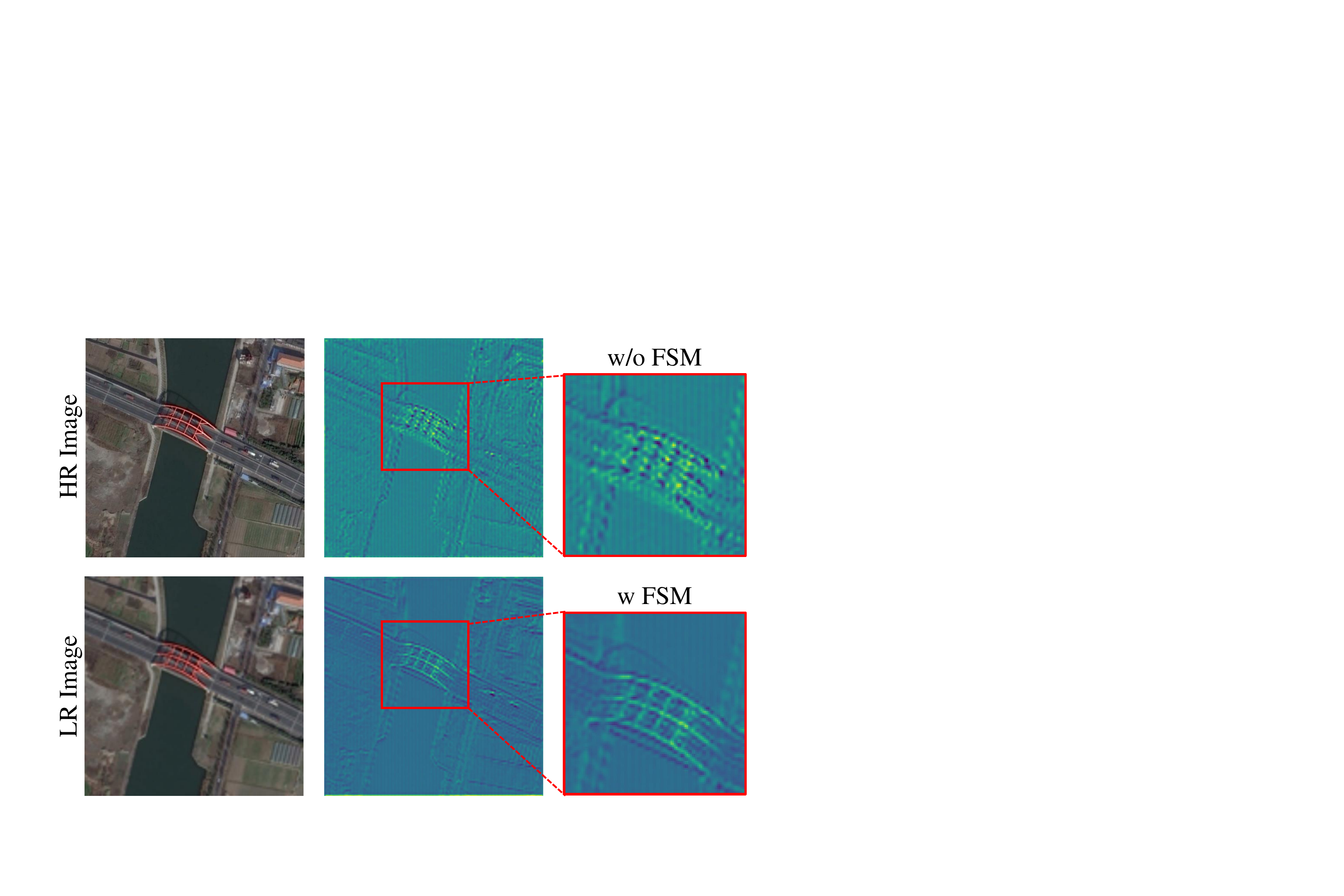}%
\captionsetup{font={scriptsize}}   
\caption{Visualization of the feature maps. The proposed Frequency Selection Module (FSM) yields sharp and clear details for reconstruction.}
\label{fre-vis}
\end{figure}

\begin{table}[!t]
  \centering
  \captionsetup{font={scriptsize}, labelsep = newline, justification=centering}
  \caption{Ablation studies for the number of FMB. The best and second-best PSNR performances are highlighted in \textbf{bold} and \underline{underline}, respectively.}
    \begin{tabular}{cccc}
    \toprule[1.5pt]
    Number of FMB & PSNR (dB)  & \#Param. & GFLOPs \\
    \midrule    
    \midrule
    2     &  27.744     & 4.55M  & 58.09G \\
    4     &   28.026    & 8.15M  & 93.18G \\
    6     &  \underline{28.178}     & 11.76M & 128.27G \\
    8     &   \textbf{28.181}    & 15.37M & 163.36G \\
    \bottomrule[1.5pt]
    \end{tabular}%
  \label{abla-num-fmb}%
\end{table}%

\begin{table}[!t]
  \centering
  \captionsetup{font={scriptsize}, labelsep = newline, justification=centering}
  \caption{Ablation studies for the number of FMG. The best and second-best PSNR performances are highlighted in \textbf{bold} and \underline{underline}, respectively.}
    \begin{tabular}{ccccc}
    \toprule[1.5pt]
    Group & 2     & 4     & 6     & 8 \\
    \midrule    
    \midrule
    \#Param. & 4.21M  & 7.99M  & 11.76M & 15.54M \\
    GFLOPs & 52.66G & 90.46G & 128.27G & 166.07G \\
    PSNR (dB)  & 27.691      & 27.985      & \textbf{28.178}      & \underline{28.154} \\
    \bottomrule[1.5pt]
    \end{tabular}%
  \label{abla-num-group}%
\end{table}%

\begin{figure}[!t]
\centering
\includegraphics[width=3.5in]{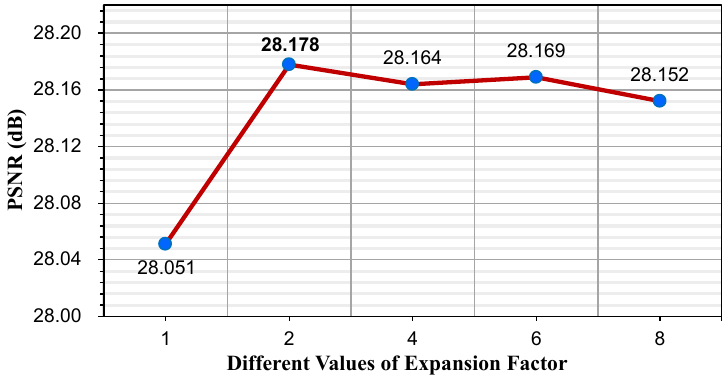}%
\captionsetup{font={scriptsize}}   
\caption{Ablation studies on the effect of different expansion factors $\lambda$.}
\label{lamda}
\end{figure}

\begin{figure}[!t]
\centering
\includegraphics[width=3.5in]{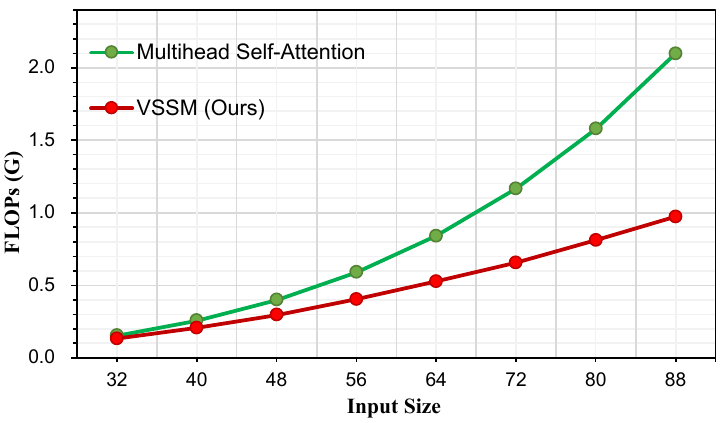}%
\captionsetup{font={scriptsize}}   
\caption{Computational complexity comparison with inputs of different resolutions. We adopt the standard Multihead Self-Attention [1] as the baseline. Initially, we adjust the model to ensure that the consumption of parameters and FLOPs is roughly equivalent. Subsequently, we increase the input resolution from $32\times32$ to $88\times88$.}
\label{inputsize}
\end{figure}

\begin{figure*}[!t]
\centering
\includegraphics[width=7in]{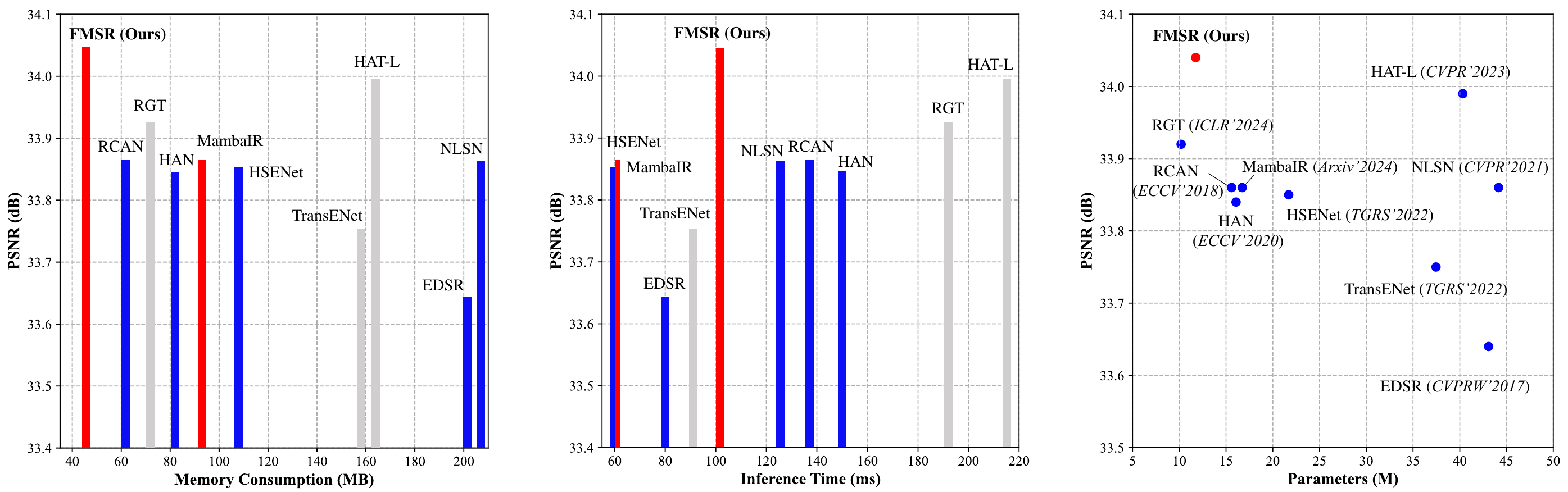}%
\captionsetup{font={scriptsize}}   
\caption{Ablation studies of memory consumption, inference times, parameters, and PSNR performance on DOTA \cite{dota}. Note that the inference times are calculated on 100 images.}
\label{efficiency}
\end{figure*}

\begin{table*}[!t]
  \centering
  \captionsetup{font={scriptsize}, labelsep = newline, justification=centering}
  \caption{Model efficient comparison. The FLOPs results are calculated with an input tensor of size $1\times3\times128\times128$. The inference times are tested on 100 random images.}
    \setlength{\tabcolsep}{1mm}{
    \begin{tabular}{ccccccccccc}
    \toprule[1.5pt]
    Metrics & EDSR \cite{edsr}  & RCAN \cite{rcan}  & HAN \cite{han}   & HSENet \cite{hsenet} & NLSN \cite{nlsa}  & TransENet \cite{transenet} & HAT-L \cite{hat} & RGT \cite{rgt}   & MambaIR \cite{mambair} & FMSR \\
    \midrule
    \midrule
    Parameters (M) & 43.09 & 15.59 & 16.07 & 21.7  & 44.15 & 37.46 & 40.32 & 10.19 & 16.72 & 11.76 \\
    FLOPs (G) & 823.34 & 261.01 & 268.89 & 306.31 & 840.79 & 87.85 & 672.15 & 40.12 & 280.82 & 128.27 \\
    Memory (MB) & 202   & 62    & 82    & 108   & 206   & 158   & 164   & 72    & 92    & 46 \\
    Inference Times (ms) & 79.188 & 136.71 & 149.24 & 59.95 & 125.42 & 91.09 & 214.56 & 191.48 & 60.99 & 100.88 \\
    \bottomrule[1.5pt]
    \end{tabular}%
    }
  \label{eff-res}%
\end{table*}%

\begin{table*}[!t]
  \centering
  \captionsetup{font={scriptsize}, labelsep = newline, justification=centering}
  \caption{Quantitative comparison on AID \cite{aid}, DOTA \cite{dota}, and DIOR \cite{dior} test set in terms of PSNR, SSIM, and LPIPS, where the 1st, 2nd, and 3rd best performance are highlighted in \textcolor{red}{\textbf{red}}, \textcolor{blue}{\textbf{blue}}, and \textcolor{green}{\textbf{green}}, respectively. FMSR++ means the self-embedding results of our FMSR.}
  \renewcommand{\arraystretch}{1.2}
  \setlength{\tabcolsep}{1mm}{
    \begin{tabular}{c|c|ccccccccc|ccc}
    \bottomrule[1.5pt]
    \multirow{2}{*}{Methods} & \multirow{2}{*}{Venue} & \multicolumn{3}{c}{AID \cite{aid}} & \multicolumn{3}{c}{DOTA \cite{dota}} & \multicolumn{3}{c|}{DIOR \cite{dior}} & \multicolumn{3}{c}{Average} \\
    \cline{3-14}
          &       & PSNR $\uparrow$  & SSIM $\uparrow$  & LPIPS $\downarrow$ & PSNR $\uparrow$  & SSIM $\uparrow$  & LPIPS $\downarrow$ & PSNR $\uparrow$  & SSIM $\uparrow$  & LPIPS $\downarrow$ & PSNR $\uparrow$      & SSIM $\uparrow$ & LPIPS $\downarrow$ \\
    \hline
    \hline
    Bicubic & -     & 28.86 & 0.7382 & 0.4803  & 31.16 & 0.7947 & 0.4043
      & 28.57 & 0.7432 & 0.4678      &     29.53  & 0.7587  & 0.4508 \\
    EDSR \cite{edsr}  & \emph{CVPRW'2017} & 30.65 & 0.8086 & 0.3068
      & 33.64 & 0.8648 & 0.2616      & 30.63 & 0.8116 & 0.3020      & 31.64 & 0.8283 & 0.2901 \\
    RDN \cite{rdn}  & \emph{CVPR'2018} & 30.74 & 0.8112 & 0.3157
      & 33.60 & 0.8670 & 0.2421      & 30.78 & 0.8147 & 0.3093      & 31.71 & 0.8310 & 0.2890\\
    RCAN \cite{rcan}  & \emph{ECCV'2018} & 30.82 & 0.8121 & 0.3112
      & 33.86 & 0.8680 & 0.2384      & 30.85 & 0.8159 & 0.3048      & 31.84 & 0.8320 & 0.2848\\
    HAN \cite{han}   & \emph{ECCV'2020} & 30.80 & 0.8122 & 0.3079
      & 33.84 & 0.8682 & 0.2363      & 30.84 & 0.8163 & 0.3026      & 31.83 & 0.8322 & 0.2823\\
    HSENet \cite{hsenet} & \emph{TGRS'2022} & 30.72 & 0.8108 &  0.3124
     & 33.85 & 0.8667 & 0.2412      & 30.77 & 0.8143 & 0.3070      & 31.78  & 0.8306 & 0.2869\\
    NLSN \cite{nlsa}  & \emph{CVPR'2021} & 30.81 & 0.8126 & 0.3044
      & 33.86 & 0.8682 & 0.2349      & 30.82 & 0.8156 & 0.3004      & 31.83 & 0.8321 & 0.2799\\
    HAUNet \cite{haunet} & \emph{TGRS'2023} & 30.88 & 0.8132 & 0.3111
      & 33.94 & 0.8687 & 0.2403      & 30.87 & 0.8160 & 0.3074      & 31.90  & 0.8326 & 0.2863\\
    TransENet \cite{transenet} & \emph{TGRS'2022} & 30.80  & 0.8109 & 0.3136
      & 33.75 & 0.8675 & 0.2418      & 30.85 & 0.8148 & 0.3089      & 31.80 & 0.8311 & 0.2881\\
    HAT-L \cite{hat} & \emph{CVPR'2023} & 30.81 & 0.8124 & 0.3078
      & \textcolor{green}{\textbf{33.99}} & 0.8684 & 0.2392     & 30.87 & 0.8161 & 0.3062      & 31.89 & 0.8323 & 0.2844 \\
    GRL-L \cite{grl}   & \emph{CVPR'2023} & 30.86  & 0.8127  & 0.3085
      & 33.86 & \textcolor{blue}{\textbf{0.8710}} &  0.2372     & 30.90 & 0.8177 &  0.3047     & 31.87 & 0.8338 & 0.2835 \\
    RGT \cite{rgt}   & \emph{ICLR'2024} & \textcolor{green}{\textbf{30.91}}  & \textcolor{blue}{\textbf{0.8159}}  & \textcolor{blue}{\textbf{0.3023}}
      & 33.92 & \textcolor{green}{\textbf{0.8709}} & \textcolor{green}{\textbf{0.2337}}      & \textcolor{green}{\textbf{30.91}} & \textcolor{blue}{\textbf{0.8182}} &  \textcolor{blue}{\textbf{0.2992}}     & \textcolor{green}{\textbf{31.91}} & \textcolor{green}{\textbf{0.8350}} & \textcolor{green}{\textbf{0.2784}}\\
    MambaIR \cite{mambair} & \emph{Arxiv'2024} & 30.85  & 0.8130  & 0.3098
      & 33.86 & 0.8691 & 0.2388      & 30.89 & 0.8167 & 0.3060      & 31.87 & 0.8329 & 0.2849\\
    \hline
    \hline
    \rowcolor{mygray}\textbf{FMSR (Ours)} &  \emph{TMM'2024}     & \textcolor{blue}{\textbf{30.93}}  & \textcolor{green}{\textbf{0.8156}}  & \textcolor{red}{\textbf{0.3001}}       & \textcolor{blue}{\textbf{34.04}} & \textcolor{blue}{\textbf{0.8710}} & \textcolor{red}{\textbf{0.2325}} & \textcolor{blue}{\textbf{30.97}} & \textcolor{blue}{\textbf{0.8187}} & \textcolor{red}{\textbf{0.2983}}      & \textcolor{blue}{\textbf{31.98}}      & \textcolor{blue}{\textbf{0.8351}} & \textcolor{red}{\textbf{0.2770}}\\
    \rowcolor{mygray}\textbf{FMSR++ (Ours)} &  \emph{TMM'2024}     & \textcolor{red}{\textbf{31.07}}  & \textcolor{red}{\textbf{0.8185}}  &  \textcolor{green}{\textbf{0.3067}}     & \textcolor{red}{\textbf{34.27}} & \textcolor{red}{\textbf{0.8735}} &    \textcolor{green}{\textbf{0.2363}}   & \textcolor{red}{\textbf{31.13}} & \textcolor{red}{\textbf{0.8219}} &     \textcolor{green}{\textbf{0.3039}}   & \textcolor{red}{\textbf{32.16}}      & \textcolor{red}{\textbf{0.8380}} & \textcolor{blue}{\textbf{0.2823}}\\
    \toprule[1.5pt]
    \end{tabular}%
    }
  \label{all-res}%
\end{table*}%

\begin{figure*}[!t]
\centering
\includegraphics[width=7in]{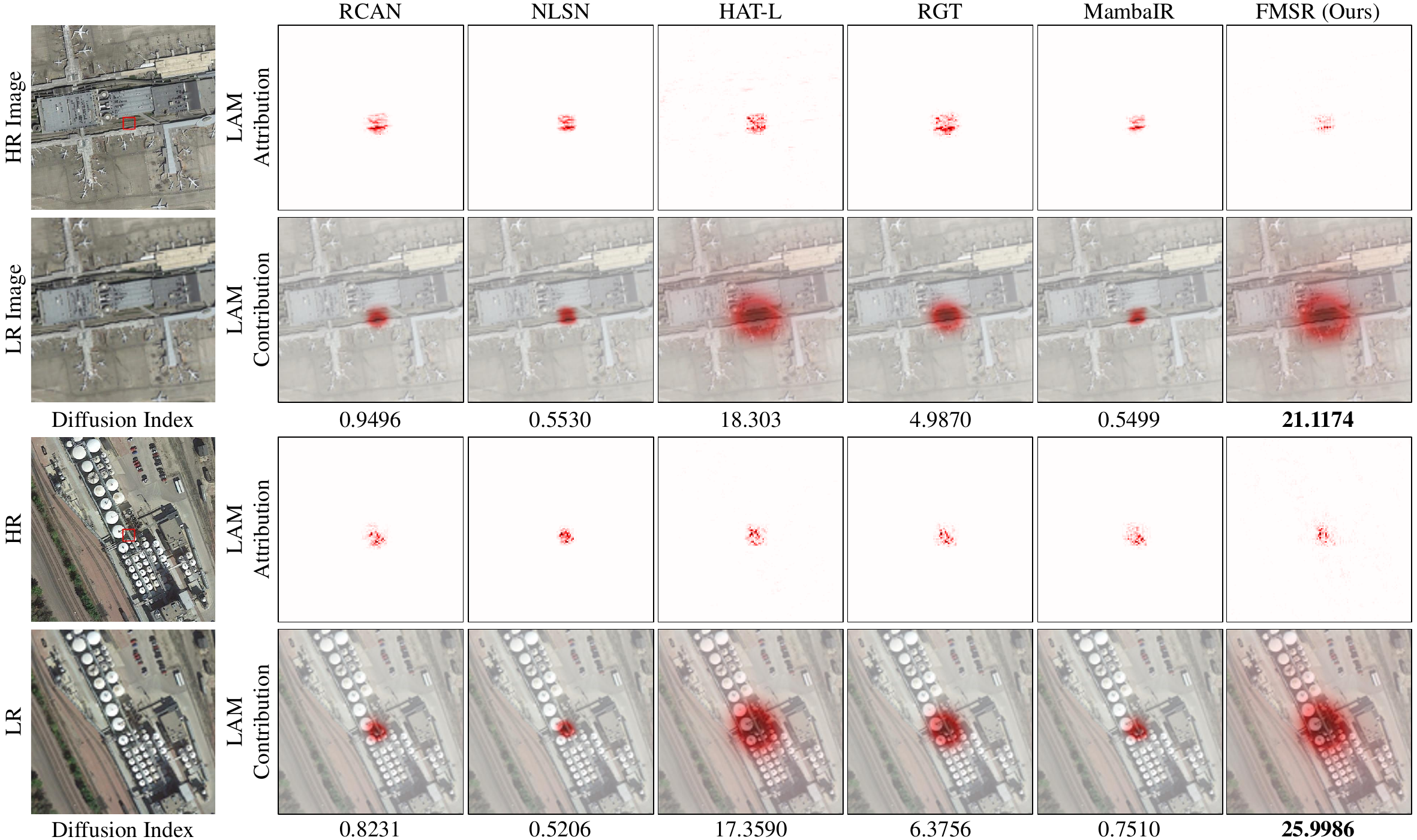}%
\captionsetup{font={scriptsize}}   
\caption{The visualization of Local Attribution Maps (LAM) \cite{lam}. A wider range of LAM illustrates more pixels are involved in reconstruction. A higher diffusion index demonstrates better global activation capability.} 
\label{lam}
\end{figure*}

\begin{table*}[!t]
  \centering
  \captionsetup{font={scriptsize}, labelsep = newline, justification=centering}
  \caption{Quantitative comparison with SOTA CNN-, Transformer-, and Mamba-based SR methods across 30 scene categories on AID \cite{aid}, where the best and second best PSNR/SSIM performance are highlighted in \textcolor{red}{\textbf{red}} and \textcolor{blue}{\textbf{blue}}, respectively.}
  \scalebox{0.92}{
    \setlength{\tabcolsep}{1.2mm}{  
  \begin{tabular}{c|cccccccccccccccc}
    \toprule[1.5pt]
    \multirow{2}[4]{*}{Categoties} & \multicolumn{2}{c}{Bicubic} & \multicolumn{2}{c}{EDSR \cite{edsr}} & \multicolumn{2}{c}{HSENet \cite{hsenet}} & \multicolumn{2}{c}{NLSN \cite{nlsa}} & \multicolumn{2}{c}{HAT-L \cite{hat}} & \multicolumn{2}{c}{MambaIR \cite{mambair}} & \multicolumn{2}{c}{RGT \cite{rgt}} & \multicolumn{2}{c}{\textbf{FMSR (Ours)}} \\
\cmidrule{2-17}          & PSNR$\uparrow$  & SSIM$\uparrow$  & PSNR$\uparrow$  & SSIM$\uparrow$  & PSNR$\uparrow$  & SSIM$\uparrow$  & PSNR$\uparrow$  & SSIM$\uparrow$  & PSNR$\uparrow$  & SSIM$\uparrow$  & PSNR$\uparrow$  & SSIM$\uparrow$  & PSNR$\uparrow$  & SSIM$\uparrow$ & PSNR$\uparrow$  & SSIM$\uparrow$\\
\midrule
\midrule
    Airport & 27.83  & 0.7554  & 29.93  & 0.8282  & 30.08  & 0.8303  & 30.16  & 0.8322  & 30.15  & 0.8319  & 30.21  & 0.8327  & \textcolor{red}{\textbf{30.29}}  & \textcolor{red}{\textbf{0.8352}}  & \textcolor{blue}{\textbf{30.26}}  & \textcolor{blue}{\textbf{0.8346}}  \\
    Bare Land & 35.60  & 0.8564  & 36.94  & 0.8837  & 36.79  & 0.8841  & \textcolor{blue}{\textbf{37.00}}  & 0.8845  & 36.88  & 0.8841  & 36.95  & 0.8843  & 37.00  & \textcolor{red}{\textbf{0.8852}}  & \textcolor{red}{\textbf{37.02}}  & \textcolor{blue}{\textbf{0.8847}}  \\
    Baseball Field & 31.00  & 0.8305  & 33.05  & 0.8765  & 33.15  & 0.8774  & 33.24  & 0.8787  & 33.25  & 0.8789  & 33.32  & 0.8792  & \textcolor{blue}{\textbf{33.35}}  & \textcolor{red}{\textbf{0.8805}}  & \textcolor{red}{\textbf{33.43}}  & \textcolor{blue}{\textbf{0.8803}}  \\
    Beach & 32.90  & 0.8446  & 34.18  & 0.8727  & 34.14  & 0.8746  & 34.31  & 0.8749  & 34.34  & 0.8756  & 34.36  & 0.8754  & \textcolor{red}{\textbf{34.41}}  & \textcolor{red}{\textbf{0.8766}}  & \textcolor{blue}{\textbf{34.38}}  & \textcolor{blue}{\textbf{0.8759}}  \\
    Bridge & 30.22  & 0.8283  & 32.93  & 0.8800  & 33.06  & 0.8809  & 33.12  & 0.8818  & 33.04  & 0.8809  & 33.14  & 0.8821  & \textcolor{blue}{\textbf{33.22}}  & \textcolor{blue}{\textbf{0.8832}}  & \textcolor{red}{\textbf{33.28}}  & \textcolor{red}{\textbf{0.8836}}  \\
    Center & 26.51  & 0.6944  & 28.77  & 0.7921  & 28.83  & 0.7937  & 28.95  & 0.7971  & 28.92  & 0.7956  & 28.99  & 0.7966  & \textcolor{blue}{\textbf{29.09}}  & \textcolor{red}{\textbf{0.8015}}  & \textcolor{red}{\textbf{29.09}}  & \textcolor{blue}{\textbf{0.8010}}  \\
    Church & 24.29  & 0.6333  & 26.30  & 0.7469  & 26.47  & 0.7507  & 26.51  & 0.7528  & 26.56  & 0.7532  & 26.61  & 0.7543  & \textcolor{red}{\textbf{26.66}}  & \textcolor{red}{\textbf{0.7580}}  & \textcolor{blue}{\textbf{26.64}}  & \textcolor{blue}{\textbf{0.7567}}  \\
    Commercial & 27.33  & 0.7174  & 29.01  & 0.7940  & 29.21  & 0.7989  & 29.21  & 0.7996  & 29.21  & 0.8007  & 29.29  & 0.8012  & \textcolor{blue}{\textbf{29.30}}  & \textcolor{red}{\textbf{0.8048}}  & \textcolor{red}{\textbf{29.34}}  & \textcolor{blue}{\textbf{0.8039}}  \\
    D-Residential & 22.93  & 0.5671  & 24.38  & 0.6839  & 24.60  & 0.6912  & 24.60  & 0.6936  & 24.67  & 0.6936  & 24.74  & 0.6965  & \textcolor{blue}{\textbf{24.78}}  & \textcolor{blue}{\textbf{0.7019}}  & \textcolor{red}{\textbf{24.78}}  & \textcolor{red}{\textbf{0.7019}}  \\
    Desert & 39.26  & 0.9100  & 40.20  & 0.9268  & 39.57  & 0.9271  & 40.27  & 0.9278  & \textcolor{red}{\textbf{40.37}}  & \textcolor{blue}{\textbf{0.9278}}  & 40.09  & 0.9270  & 40.15  & \textcolor{red}{\textbf{0.9283}}  & \textcolor{blue}{\textbf{40.29}}  & 0.9275  \\
    Farmland & 33.10  & 0.8226  & 35.00  & 0.8683  & 35.02  & 0.8692  & 35.10  & 0.8699  & 35.03  & 0.8691  & 35.09  & 0.8696  & \textcolor{blue}{\textbf{35.14}}  & \textcolor{red}{\textbf{0.8715}}  & \textcolor{red}{\textbf{35.18}}  & \textcolor{blue}{\textbf{0.8711}}  \\
    Forest & 28.79  & 0.6605  & 29.85  & 0.7315  & 30.00  & 0.7363  & 29.98  & 0.7369  & \textcolor{blue}{\textbf{30.01}}  & 0.7363  & 30.00  & 0.7370  & 30.01  & \textcolor{red}{\textbf{0.7400}}  & \textcolor{red}{\textbf{30.04}}  & \textcolor{blue}{\textbf{0.7391}}  \\
    Industrial & 26.77  & 0.6952  & 28.88  & 0.7931  & 28.98  & 0.7956  & 29.04  & 0.7982  & 29.04  & 0.7980  & 29.09  & 0.7988  & \textcolor{red}{\textbf{29.20}}  & \textcolor{red}{\textbf{0.8034}}  & \textcolor{blue}{\textbf{29.17}}  & \textcolor{blue}{\textbf{0.8027}}  \\
    Meadow & 33.86  & 0.7483  & 34.63  & 0.7804  & 34.55  & 0.7804  & 34.69  & 0.7821  & \textcolor{blue}{\textbf{34.70}}  & 0.7815  & 34.64  & 0.7811  & 34.69  & \textcolor{red}{\textbf{0.7831}}  & \textcolor{red}{\textbf{34.74}}  & \textcolor{blue}{\textbf{0.7827}}  \\
    M-Residential & 26.36  & 0.6335  & 28.34  & 0.7365  & 28.45  & 0.7390  & 28.49  & 0.7418  & 28.46  & 0.7408  & 28.59  & 0.7435  & \textcolor{red}{\textbf{28.66}}  & \textcolor{red}{\textbf{0.7479}}  & \textcolor{blue}{\textbf{28.65}}  & \textcolor{blue}{\textbf{0.7472}}  \\
    Mountain & 29.51  & 0.7349  & 30.63  & 0.7885  & 30.72  & 0.7907  & 30.74  & 0.7916  & \textcolor{blue}{\textbf{30.78}}  & 0.7923  & 30.76  & 0.7918  & 30.71  & \textcolor{red}{\textbf{0.7941}}  & \textcolor{red}{\textbf{30.78}}  & \textcolor{blue}{\textbf{0.7928}}  \\
    Park  & 29.06  & 0.7530  & 30.54  & 0.8130  & 30.71  & 0.8167  & 30.71  & 0.8177  & 30.71  & 0.8189  & 30.76  & 0.8185  & \textcolor{red}{\textbf{30.81}}  & \textcolor{red}{\textbf{0.8211}}  & \textcolor{blue}{\textbf{30.80}}  & \textcolor{blue}{\textbf{0.8204}}  \\
    Parking & 24.24  & 0.7060  & 27.25  & 0.8317  & 27.32  & 0.8341  & 27.57  & 0.8408  & 27.56  & 0.8405  & 27.65  & 0.8414  & \textcolor{blue}{\textbf{27.79}}  & \textcolor{blue}{\textbf{0.8472}}  & \textcolor{red}{\textbf{28.02}}  & \textcolor{red}{\textbf{0.8524}}  \\
    Playground & 32.64  & 0.8450  & 35.37  & 0.8943  & 35.46  & 0.8952  & 35.58  & 0.8967  & 35.49  & 0.8959  & 35.54  & 0.8961  & \textcolor{blue}{\textbf{35.65}}  & \textcolor{blue}{\textbf{0.8978}}  & \textcolor{red}{\textbf{35.72}}  & \textcolor{red}{\textbf{0.8978}}  \\
    Pond  & 30.70  & 0.8167  & 32.11  & 0.8542  & 32.17  & 0.8549  & 32.22  & 0.8559  & 32.18  & 0.8555  & \textcolor{blue}{\textbf{32.24}}  & 0.8559  & \textcolor{red}{\textbf{32.24}}  & \textcolor{red}{\textbf{0.8567}}  & 32.23  & \textcolor{blue}{\textbf{0.8560}}  \\
    Port  & 26.67  & 0.7986  & 28.50  & 0.8596  & 28.71  & 0.8623  & 28.71  & 0.8631  & 28.81  & 0.8638  & 28.83  & 0.8644  & \textcolor{blue}{\textbf{28.90}}  & \textcolor{blue}{\textbf{0.8667}}  & \textcolor{red}{\textbf{28.92}}  & \textcolor{red}{\textbf{0.8668}}  \\
    Railway Station & 26.78  & 0.6793  & 28.72  & 0.7738  & 28.84  & 0.7762  & 28.89  & 0.7783  & 28.88  & 0.7780  & 28.99  & 0.7802  & \textcolor{blue}{\textbf{29.03}}  & \textcolor{red}{\textbf{0.7833}}  & \textcolor{red}{\textbf{29.04}}  & \textcolor{blue}{\textbf{0.7830}}  \\
    Resort & 26.79  & 0.7029  & 28.52  & 0.7799  & 28.64  & 0.7825  & 28.68  & 0.7845  & 28.71  & 0.7849  & 28.76  & 0.7857  & \textcolor{blue}{\textbf{28.80}}  & \textcolor{red}{\textbf{0.7889}}  & \textcolor{red}{\textbf{28.82}}  & \textcolor{blue}{\textbf{0.7886}}  \\
    River & 30.37  & 0.7402  & 31.55  & 0.7891  & 31.61  & 0.7904  & 31.64  & 0.7914  & 31.63  & 0.7909  & 31.64  & 0.7912  & \textcolor{red}{\textbf{31.67}}  & \textcolor{red}{\textbf{0.7927}}  & \textcolor{blue}{\textbf{31.66}}  & \textcolor{blue}{\textbf{0.7918}}  \\
    School & 27.41  & 0.7237  & 29.36  & 0.8044  & 29.51  & 0.8074  & 29.55  & 0.8097  & 29.54  & 0.8104  & 29.62  & 0.8113  & \textcolor{red}{\textbf{29.71}}  & \textcolor{red}{\textbf{0.8150}}  & \textcolor{blue}{\textbf{29.70}}  & \textcolor{blue}{\textbf{0.8143}}  \\
    S-Residential & 26.66  & 0.6006  & 27.71  & 0.6728  & 27.84  & 0.6754  & 27.84  & 0.6767  & 27.88  & 0.6759  & 27.88  & 0.6768  & \textcolor{blue}{\textbf{27.91}}  & \textcolor{red}{\textbf{0.6796}}  & \textcolor{red}{\textbf{27.91}}  & \textcolor{blue}{\textbf{0.6789}}  \\
    Square & 28.55  & 0.7391  & 30.84  & 0.8200  & 30.94  & 0.8223  & 31.04  & 0.8244  & 31.00  & 0.8251  & 31.05  & 0.8247  & \textcolor{blue}{\textbf{31.16}}  & \textcolor{red}{\textbf{0.8284}}  & \textcolor{red}{\textbf{31.16}}  & \textcolor{blue}{\textbf{0.8276}}  \\
    Stadium & 27.16  & 0.7547  & 29.63  & 0.8387  & 29.68  & 0.8391  & 29.79  & 0.8422  & 29.77  & 0.8422  & 29.83  & 0.8421  & \textcolor{blue}{\textbf{29.94}}  & \textcolor{blue}{\textbf{0.8457}}  & \textcolor{red}{\textbf{29.97}}  & \textcolor{red}{\textbf{0.8460}}  \\
    Storage Tanks & 25.65  & 0.6793  & 27.44  & 0.7664  & 27.58  & 0.7688  & 27.61  & 0.7709  & 27.60  & 0.7698  & 27.64  & 0.7709  & \textcolor{blue}{\textbf{27.71}}  & \textcolor{red}{\textbf{0.7744}}  & \textcolor{red}{\textbf{27.71}}  & \textcolor{blue}{\textbf{0.7739}}  \\
    Viaduct & 26.97  & 0.6755  & 28.99  & 0.7757  & 29.08  & 0.7772  & 29.17  & 0.7813  & 29.11  & 0.7794  & 29.19  & 0.7810  & \textcolor{blue}{\textbf{29.26}}  & \textcolor{red}{\textbf{0.7851}}  & \textcolor{red}{\textbf{29.26}}  & \textcolor{blue}{\textbf{0.7840}}  \\
\midrule
\midrule
    Average & 28.86  & 0.7382  & 30.65  & 0.8086  & 30.72  & 0.8108  & 30.81  & 0.8126  & 30.81  & 0.8124  & 30.85  & 0.8130  & \textcolor{blue}{\textbf{30.91}}  & \textcolor{red}{\textbf{0.8159}}  & \textcolor{red}{\textbf{30.93}}  & \textcolor{blue}{\textbf{0.8156}}  \\
    \bottomrule[1.5pt]
    \end{tabular}%
    }}
  \label{aid-res}%
\end{table*}%

\begin{figure*}[ht]
\centering
\includegraphics[width=7in]{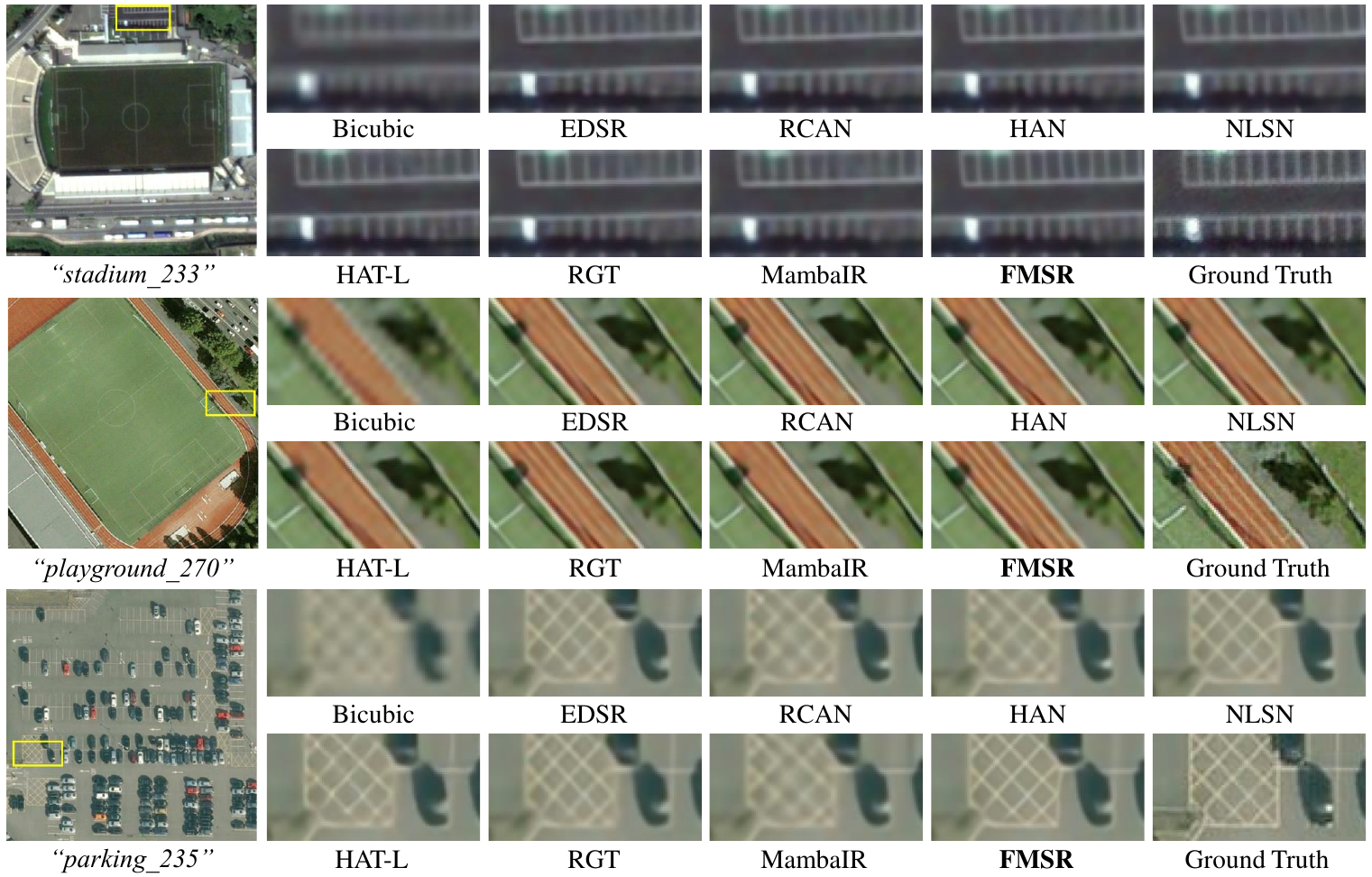}%
\captionsetup{font={scriptsize}}   
\caption{Visual comparisons of our FMSR with CNN-, Transformer-, and Mamba-based methods on AID \cite{aid} with scale $\times4$. Zoom in for better observation.}
\label{vis-aid}
\end{figure*}

\begin{figure*}[!t]
\centering
\includegraphics[width=7in]{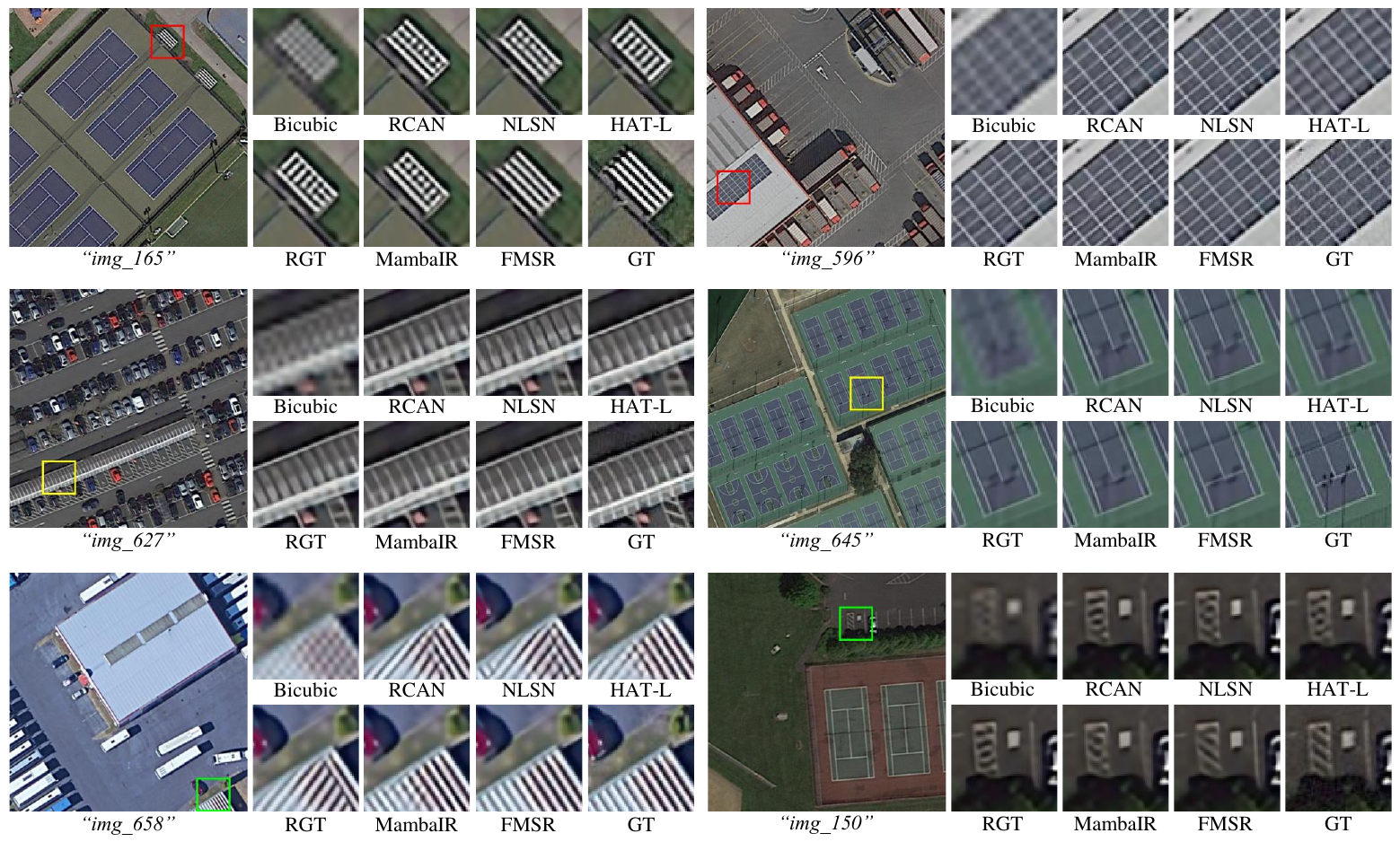}%
\captionsetup{font={scriptsize}}   
\caption{Visual comparisons of our FMSR with CNN-, Transformer-, and Mamba-based methods on DOTA \cite{dota} with scale $\times4$. Zoom in for better observation.}
\label{vis-dota}
\end{figure*}

\section{Experiment}\label{section4}
\subsection{Datasets}
In this paper, we report the results of the SR performance on three RSI benchmarks, including AID \cite{aid}, DOTA \cite{dota}, and DIOR \cite{dior}. In particular, AID is used to form the training and test simultaneously. We randomly select 3000 and 900 images from AID to form the training and test set, with an image size of $640\times640$. Note that the training and test parts of AID are non-overlapping. In DOTA and DIOR, 900 and 1000 images are randomly extracted for model evaluation, respectively. Both of them are with a size of $512\times512$.

\subsection{Implementation Details}
\textbf{Model Details}. This paper focuses on $\times4$ SR. Our FMSR is constructed by 6 FMGs for deep feature exploration, with each FMG consisting of 6 FMBs, \emph{i.e.,} $m=n=6$. Empirically, we set the internal channel dimension to $c=96$. ll convolutional kernel sizes are set to $3\times3$, except for those in the Hybrid Gate Module (HGM) and Frequency Selection Module (FSM), which utilize $1\times1$ kernels for increased efficiency. The expansion rate in the linear projection layer is set to $\lambda=2$.

\textbf{Training Details}. During the training procedure, all the SR methods were retrained on the AID training set using L1 loss and ADAM algorithm with $\beta_1=0.9$ and $\beta_2=0.999$. To train our FMSR, we randomly select 4 image patches with the size of $64\times64$ in each mini-batch. The learning rate is initialized to $1\times10^{-4}$ and halved every 200 epochs until training stops at 500 epochs. All SR models were implemented in the PyTorch framework and trained on a single NVIDIA RTX 3090 GPU with 24 GB memory and a 3.40 GHz AMD Ryzen 5700X CPU.

\subsection{Evaluation Metrics}
Two classical full-reference indicators are used to evaluate the SR performance, \emph{i.e.,} Peak Signal-to-Noise Ratio (PSNR) and Structural Similarity Index (SSIM) \cite{ssim}. Moreover, the LPIPS metric \cite{lpips} is employed to analyze the perceptual quality of super-resolved results. Note that PSNR and SSIM are calculated on the luminance channel (Y) of YCbCr space.

\subsection{Ablation Study}
In this section, we discuss the proposed FMSR in depth by investigating the effect of its major components and their variants. All these models are trained on AID with scale factor $\times4$. Following prior work \cite{ttst}, we employ a small-scale dataset, AID-tiny, consisting of 30 randomly selected images from AID, for efficient evaluation unless otherwise specified. The baseline model is derived by excluding FSM and substituting VVSM and HGM with the standard window-based self-attention and 5 residual blocks, respectively.

\subsubsection{Effect of Key Modules} \textbf{(a) Effect of VSSM.} Table \ref{tb1} reports that the Model-1 (baseline) obtains 27.751 dB. Model-2 exhibits a performance gain of 0.089 dB over the baseline model, which demonstrates the effectiveness of VSSM in global modeling. To demonstrate the linear complexity of VSSM, we conducted complexity experiments with inputs of different resolutions, as shown in Fig. \ref{inputsize}. Specifically, we adopted the standard Multihead Self-Attention (MSA) \cite{vaswani2017attention} with a dimension of 180 as the baseline and adjusted our model to have a dimension of 144. This adjustment ensured that the complexity in terms of parameters (0.1308M vs. 0.1279M) and FLOPs (0.1534G vs. 0.1312G) was roughly equivalent. Our method is more efficient than the widely used MSA and exhibits linear complexity with input resolution. In addition, in Fig. \ref{fea-vis}, we visualize the intermediate feature maps of the baseline model and our FMSR. The features are obtained by visualizing the 56th channels at the end of FMG. With VSSM, the results of our FMSR are more prominent across the entire feature maps, highlighting the favorable global modeling capability.

\textbf{(b) Effect of HGM.} Employing both VSSM and the MLP layer, Model-3 demonstrates a favorable performance improvement of 0.258 dB. However, when equipped with the channel attention mechanism to introduce more locality, Model-4 performs similarly to Model-2 (28.088 dB vs. 28.104 dB). This suggests that neither MLP nor channel attention is less effective in boosting the performance of Mamba. In this context, further improving the reconstruction performance becomes very challenging. Our HGM achieves a performance gain of 0.034 dB. Thus, to benefit from both global and local inductive bias, we combine VSSM and HGM in our FMSR. In addition, we have investigated the impact of different expansion factors $\lambda$ used in the linear projection layer. The quantitative results are presented in Fig. \ref{lamda}. It is observed that the performance of FMSR does not fluctuate significantly with changes in $\lambda$. We ultimately selected $\lambda=2$ as it delivers the best performance.

\textbf{(c) Effect of FSM.} By comparing the results of FMSR and Model-5, we can evaluate the performance of the proposed FSM. In this case, our FSM can bring an improvement of 0.056 dB. As reported in Table \ref{abla-fsm}, we discuss some variants of FSM shown in Fig. \ref{fsm}. If we do not perform frequency selection, FSM(a) produces the worst PSNR performance. By inserting a simple ReLU activation, FSM(b) could adaptively eliminate noisy frequencies and achieve a gain of 0.018 dB, demonstrating the effectiveness of the selection mechanism. If we adopt $1\times1$ convolution and GELU for selection, FMSR allows for a more flexible threshold for frequency selection, thus obtaining the best performance. Moreover, visual comparisons between FMSR and Model-5 are shown in Fig. \ref{fre-vis}. In Fig. \ref{fre-vis}, our FMSR generates superior textures in the high-frequency of the bridge. In contrast, without performing frequency selection, the intermediate features are blurred at the boundary and unclear at the edge. These results confirm that our FSM has the ability to explore more critical high-frequency cues for better SR performance.

\subsubsection{Effect of HAI} We show the influence of HAI in Table \ref{abla-hai}, where we conduct three ablation analyses: 1) without HAI, 2) with residual connection (Skip), and with HAI. We observe that without HAI, the performance of FMSR drops by 0.13 dB. Additionally, comparing the model with Skip, our FMSR achieves a significant improvement of 0.36 dB. This may be because of the misalignment between global and local representations, which means simply adding them may not elaborate enough to integrate these different levels of knowledge, thus generating suboptimal performance. Benefiting from the adaptive scaling factor $\alpha$, our FMSR could dynamically adjust the features at the global and local ranges, thus generating enriched feature integration.

\subsubsection{Model Efficiency} The parameters, FLOPs, and SR performance of state-of-the-art (SOTA) methods are reported in Table \ref{all-res}. Intuitively, we plot the performance versus parameters in Fig. \ref{efficiency}(c), where we observe that FMSR strikes a favorable trade-off between performance and parameters. Here, we further investigate the relationship between network structure and model complexity. Moreover, more intuitive metrics are involved to analyze the model efficiency, such as inference times and memory consumption.

\textbf{(a) Number of FMB and FMG.} We study the inference of the number of FMB and FMG in Table \ref{abla-num-fmb} and Table \ref{abla-num-group}, respectively. As we can see in Table \ref{abla-num-fmb}, using more FMB leads to the consistently increasing PSNR performance from 27.744 to 28.181 dB, demonstrating the effectiveness of FMB. Meanwhile, the parameters grow dramatically from 4.55 to 15.37M. To strike a favorable balance between performance and model size, we finally picked 6 FMB in our FMSR. Regarding the number of FMG, as shown in Table \ref{abla-num-group}, the increasing number of groups leads to higher PSNR values. Nevertheless, the performance saturates at group 8, which may be caused by over-fitting. Ultimately, we insert 6 FMG in the FMSR for deep feature learning.

\textbf{(d) Memory Consumption.}
The max CUDA memory consumption of our FMSR and SOTA models are shown in Fig. \ref{efficiency}(a), from which we can see that our consumes the least memory consumption while outperforming other methods. Specifically, compared to competitive NLSN that adopts non-local attention for global modeling, we find the performance of FMSR is 0.18 dB higher, but also reduces the memory by 160 MB. Furthermore, FMSR achieves better performance against impressive HAT-L with only 28\% of its memory. This indicates that FMSR has stronger global modeling capability and greatly surpasses Transformer-based models in model efficiency.

\textbf{(c) Inference Times.} Regarding the inference times, as shown in Fig. \ref{efficiency}(b), our FMSR shows a trade-off with other methods. For instance, FMSR achieves the best PSNR performance compared to the advanced Transformer-based model RGT by 0.12 dB, but at the cost of increased inference time usage of about 90.6 ms, measured by the test times of 100 images. These results provide intuitive evidence that our FMSR is very efficient in large-scale remote sensing image SR tasks.

\begin{figure*}[!t]
\centering
\includegraphics[width=7in]{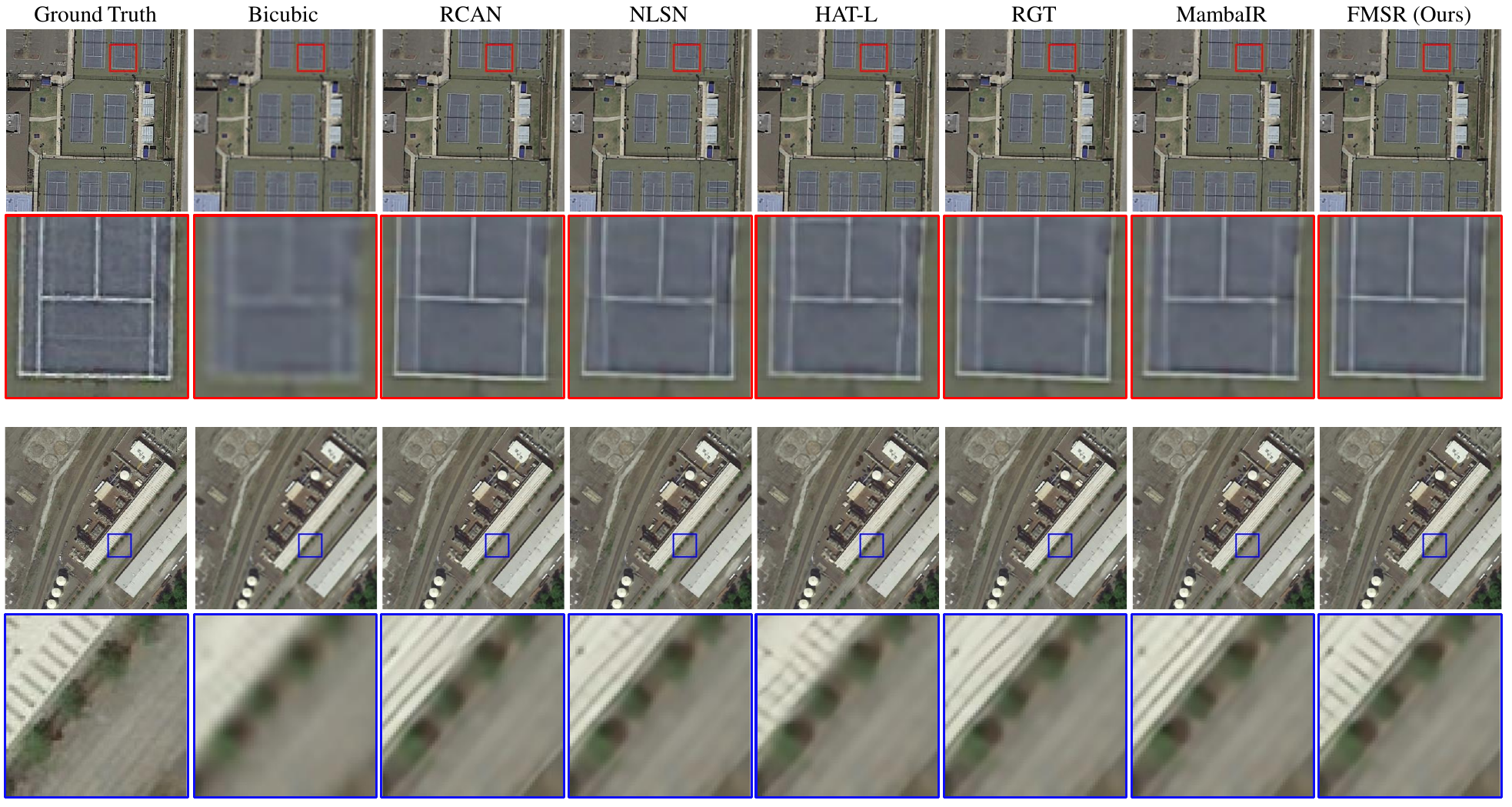}%
\captionsetup{font={scriptsize}}   
\caption{Visual comparisons of our FMSR with CNN-, Transformer-, and Mamba-based methods on DIOR \cite{dior} with scale $\times4$. Zoom in for better observation.}
\label{vis-dior}
\end{figure*}
\subsection{Comparisons With State-of-the-Art}
\subsubsection{Comparative Methods}To evaluate the SR performance of our FMSR against SOTA methods on remote sensing images, advanced CNN-, Transformer-, and Mamba-based models are involved for comprehensive comparison, including EDSR \cite{edsr}, RDN \cite{rdn}, RCAN \cite{rcan}, HAN \cite{han}, NLSN \cite{nlsa}, HSENet \cite{hsenet}, TransENet \cite{transenet}, HAT-L \cite{hat}, GRL-L \cite{grl}, RGT \cite{rgt}, and MambaIR \cite{mambair}. We also report the self-ensemble results of our FMSR, dubbed FMSR++.

\subsubsection{Quantitative Evaluations} The quantitative results on the AID, DOTA, and DIOR datasets are shown in Table. \ref{all-res}. Our FMSR++ and FMSR achieve the highest and second-highest average performance on all metrics, demonstrating their superior SR performance across various remote sensing benchmarks. Particularly on the AID dataset, FMSR generates a substantial gain of 0.12 dB PSNR over HAT-L with only 29\% parameters and 19\% FLOPs. Compared to GRL-L, FMSR obtains 0.18 dB higher PSNR on the DOTA dataset with 60\% lower complexity. Our FMSR exhibits 0.3001 LPIPS on AID, which is much higher than other comparative approaches. Moreover, recent Mamba-based methods, like MambaIR, achieved marginal improvements against CNN-based models. For instance, compared to NLSN, which explores global dependency with non-local attention, MambaIR only leads NLSN by 0.04 dB on AID. This underscores the effectiveness of VSSM in exploring non-local dependency. However, it also indicates that spatial-wise global modeling reaches a plateau in complex remote sensing images. In contrast to simply employing VSSM for global modeling, our FMSR seeks a more practical solution in the frequency domain, thus achieving significant improvement over MambaIR.

In addition, to validate the generalization capability of SR models on diverse remote sensing senses, we further report the PSNR and SSIM results on AID across 30 scene categories. The results are shown in Table \ref{aid-res}. As we can see, FMSR demonstrates stronger generalization capability against SOTA methods and obtains the best performance on almost all remote sensing scenarios, receiving 0.18 dB PSNR and 0.039 SSIM over MambaIR on the "Industrial" scene. Furthermore, in the "Stadium" category, our FMSR significantly outperforms HAT-L by a large margin. These notable improvements obtained by FMSR align with our motivation, which aims to introduce Mamba for efficient yet favorable global modeling in large-scale remote sensing images.

\subsubsection{Qualitative Results} Visual comparisons on AID, DOTA, and DIOR with a scale factor of $\times4$ are shown in Fig. \ref{vis-aid}, Fig. \ref{vis-dota}, and Fig. \ref{vis-dior}. From these visualizations, we observe that our FMSR recovers sharp edges with richer textures, especially capturing critical high-frequency details in these remote sensing images. For example, comparing the restored "playground\_270" in Fig. \ref{vis-aid}, we can see that the recent competitive Transformer-based method HAT-L and Mamba-based model MambaIR struggle to reconstruct the HR lines on the runway. Moreover, as illustrated in "img\_165" and "img\_658" of Fig. \ref{vis-dota}, only the proposed FMSR can recover severely damaged lines on the ground, while other SR models fail to produce accurate distribution on the details. These visual comparisons further demonstrate the effectiveness of our FMSR in capturing and reconstructing fine details in RSI. 

When comparing the visual results of the top image from the DIOR dataset in Fig. \ref{vis-dior}, where large-scale and global information exists, we observe that all CNN-, Transformer-, and Mamba-based SR networks struggle to handle this issue, resulting in suboptimal results and losing some high-frequency textures. In contrast, benefiting from the global modeling capability of VSSM as well as the spatial-frequency dual-domain exploration, our FMSR produces results that are visually close to the ground truth and successfully generate multi-frequency lines. For example, MambaIR without frequency selection cannot recover accurate details, while HAT-L produces severe distortion. Our FMSR still maintains favorable visual quality with more high-frequency contextual information. Similarly, the visual comparisons in the bottom image of Fig. \ref{vis-dior} provide additional evidence of the superiority of our FMSR. 

In addition, the LAM comparisons shown in Fig. \ref{lam} demonstrate that FMSR can exploit more pixels during the SR process thanks to the favorable long-range modeling capability of VSSM. By comparing FMSR with HAT-L, FMSR surpasses HAT-L by 8.637 in terms of the diffusion index, indicating the strong wide-range pixel utilization capability due to our spatial-frequency dual-domain representation.

\section{Conclusion}\label{section5}
In this study, we first introduce the state space model for remote sensing image super-resolution. FMSR effectively models global dependencies in large-scale remote sensing images while enjoying the linear complexity. Specifically, we develop an efficient yet effective frequency selection module (FSM) to incorporate more relevant frequencies during spatial-frequency dual-domain learning. Meanwhile, channel-wise attention metrics are linearly scaled to enrich the spatially varying representation. Furthermore, to combine global and local representations, we employ a learnable adaptor to adaptively adjust features across different levels. Extensive quantitative and qualitative experiments across AID, DOTA, and DIOR benchmarks demonstrate the superior performance of our FMSR in remote sensing image super-resolution tasks compared to state-of-the-art CNN-, Transformer-, and Mamba-based SR models.

As prior research demonstrated that achieving optimal performance on the $\times4$ SR task typically translates to favorable performance on lower scaling factors, this study focused solely on the $\times4$ SR. While saving computational resources, this specificity lacks flexibility in exploring SR at different scales. In future work, we plan to extend our FMSR to include more scaling factors, further demonstrating its robustness and effectiveness.



%
%


%

%


%
%

\ifCLASSOPTIONcaptionsoff
  \newpage
\fi



%
\bibliographystyle{IEEEtran}
\bibliography{reference}

\begin{thebibliography}{10}
\providecommand{\url}[1]{#1}
\csname url@samestyle\endcsname
\providecommand{\newblock}{\relax}
\providecommand{\bibinfo}[2]{#2}
\providecommand{\BIBentrySTDinterwordspacing}{\spaceskip=0pt\relax}
\providecommand{\BIBentryALTinterwordstretchfactor}{4}
\providecommand{\BIBentryALTinterwordspacing}{\spaceskip=\fontdimen2\font plus
\BIBentryALTinterwordstretchfactor\fontdimen3\font minus \fontdimen4\font\relax}
\providecommand{\BIBforeignlanguage}[2]{{%
\expandafter\ifx\csname l@#1\endcsname\relax
\typeout{** WARNING: IEEEtran.bst: No hyphenation pattern has been}%
\typeout{** loaded for the language `#1'. Using the pattern for}%
\typeout{** the default language instead.}%
\else
\language=\csname l@#1\endcsname
\fi
#2}}
\providecommand{\BIBdecl}{\relax}
\BIBdecl

\bibitem{tmm1}
J.~Zhang, Y.~Rao, X.~Huang, G.~Li, X.~Zhou, and D.~Zeng, ``Frequency-aware multi-modal fine-tuning for few-shot open-set remote sensing scene classification,'' \emph{IEEE Transactions on Multimedia}, vol.~26, pp. 7823--7837, 2024.

\bibitem{chen2024changemamba}
H.~Chen, J.~Song, C.~Han, J.~Xia, and N.~Yokoya, ``Changemamba: Remote sensing change detection with spatio-temporal state space model,'' \emph{IEEE Transactions on Geoscience and Remote Sensing}, vol.~62, pp. 1--20, 2024.

\bibitem{chen2023exchange}
H.~Chen, J.~Song, C.~Wu, B.~Du, and N.~Yokoya, ``Exchange means change: An unsupervised single-temporal change detection framework based on intra-and inter-image patch exchange,'' \emph{ISPRS Journal of Photogrammetry and Remote Sensing}, vol. 206, pp. 87--105, 2023.

\bibitem{wjj2024isprs}
J.~Wang, A.~Ma, Z.~Chen, Z.~Zheng, Y.~Wan, L.~Zhang, and Y.~Zhong, ``Earthvqanet: Multi-task visual question answering for remote sensing image understanding,'' \emph{ISPRS Journal of Photogrammetry and Remote Sensing}, vol. 212, pp. 422--439, 2024.

\bibitem{tmm3}
P.~Liu, T.~Xu, H.~Chen, S.~Zhou, H.~Qin, and J.~Li, ``Spectrum-driven mixed-frequency network for hyperspectral salient object detection,'' \emph{IEEE Transactions on Multimedia}, vol.~26, pp. 5296--5310, 2024.

\bibitem{guo2024saan}
H.~Guo, X.~Su, C.~Wu, B.~Du, and L.~Zhang, ``Saan: Similarity-aware attention flow network for change detection with vhr remote sensing images,'' \emph{IEEE Transactions on Image Processing}, vol.~33, pp. 2599--2613, 2024.

\bibitem{du2023global}
H.~Du, W.~Zhan, Z.~Liu, E.~S. Krayenhoff, L.~Zhao, L.~Jiang, P.~Dong, L.~Li, F.~Huang \emph{et~al.}, ``Global mapping of urban thermal anisotropy reveals substantial potential biases for remotely sensed urban climates,'' \emph{Science Bulletin}, vol.~68, no.~16, pp. 1809--1818, 2023.

\bibitem{liu2023rethinking}
X.~Liu, J.~Hou, X.~Cong, H.~Shen, Z.~Lou, L.-J. Deng, and J.~W. You, ``Rethinking pan-sharpening via spectral-band modulation,'' \emph{IEEE Transactions on Geoscience and Remote Sensing}, 2023.

\bibitem{zhang2023hyperspectral}
Q.~Zhang, Y.~Zheng, Q.~Yuan, M.~Song, H.~Yu, and Y.~Xiao, ``Hyperspectral image denoising: From model-driven, data-driven, to model-data-driven,'' \emph{IEEE Transactions on Neural Networks and Learning Systems}, pp. 1--21, 2023.

\bibitem{chen2021hybrid}
X.~Chen, Y.~Li, L.~Dai, and C.~Kong, ``Hybrid high-resolution learning for single remote sensing satellite image dehazing,'' \emph{IEEE geoscience and remote sensing letters}, vol.~19, pp. 1--5, 2021.

\bibitem{zhang2022cooperated}
Q.~Zhang, Q.~Yuan, M.~Song, H.~Yu, and L.~Zhang, ``Cooperated spectral low-rankness prior and deep spatial prior for hsi unsupervised denoising,'' \emph{IEEE Transactions on Image Processing}, vol.~31, pp. 6356--6368, 2022.

\bibitem{tmm5}
Y.~Zhao, Q.~Teng, H.~Chen, S.~Zhang, X.~He, Y.~Li, and R.~E. Sheriff, ``Activating more information in arbitrary-scale image super-resolution,'' \emph{IEEE Transactions on Multimedia}, vol.~26, pp. 7946--7961, 2024.

\bibitem{tmm6}
J.-S. Yoo, D.-W. Kim, Y.~Lu, and S.-W. Jung, ``Rzsr: Reference-based zero-shot super-resolution with depth guided self-exemplars,'' \emph{IEEE Transactions on Multimedia}, vol.~25, pp. 5972--5983, 2023.

\bibitem{xiao2020space}
Z.~Xiao, Z.~Xiong, X.~Fu, D.~Liu, and Z.-J. Zha, ``Space-time video super-resolution using temporal profiles,'' in \emph{Proceedings of the 28th ACM International Conference on Multimedia}, 2020, pp. 664--672.

\bibitem{xiao2021space}
Z.~Xiao, X.~Fu, J.~Huang, Z.~Cheng, and Z.~Xiong, ``Space-time distillation for video super-resolution,'' in \emph{Proceedings of the IEEE/CVF conference on computer vision and pattern recognition}, 2021, pp. 2113--2122.

\bibitem{kim2010single}
K.~I. Kim and Y.~Kwon, ``Single-image super-resolution using sparse regression and natural image prior,'' \emph{IEEE transactions on pattern analysis and machine intelligence}, vol.~32, no.~6, pp. 1127--1133, 2010.

\bibitem{tmm7}
J.~Jiang, X.~Ma, C.~Chen, T.~Lu, Z.~Wang, and J.~Ma, ``Single image super-resolution via locally regularized anchored neighborhood regression and nonlocal means,'' \emph{IEEE Transactions on Multimedia}, vol.~19, no.~1, pp. 15--26, 2017.

\bibitem{zhou2024PDR}
J.~Zhou, S.~Wang, Z.~Lin, Q.~Jiang, and F.~Sohel, ``A pixel distribution remapping and multi-prior retinex variational model for underwater image enhancement,'' \emph{IEEE Transactions on Multimedia}, pp. 1--12, 2024.

\bibitem{zhou2023adp}
J.~Zhou, Q.~Liu, Q.~Jiang, W.~Ren, K.-M. Lam, and W.~Zhang, ``Underwater camera: improving visual perception via adaptive dark pixel prior and color correction,'' \emph{International Journal of Computer Vision}, pp. 1--19, 2023.

\bibitem{xiao2023cutmib}
Z.~Xiao, Y.~Liu, R.~Gao, and Z.~Xiong, ``Cutmib: Boosting light field super-resolution via multi-view image blending,'' in \emph{Proceedings of the IEEE/CVF Conference on Computer Vision and Pattern Recognition}, 2023, pp. 1672--1682.

\bibitem{erf}
W.~Luo, Y.~Li, R.~Urtasun, and R.~Zemel, ``Understanding the effective receptive field in deep convolutional neural networks,'' \emph{Advances in neural information processing systems}, vol.~29, 2016.

\bibitem{nlsa}
Y.~Mei, Y.~Fan, and Y.~Zhou, ``Image super-resolution with non-local sparse attention,'' in \emph{Proceedings of the IEEE/CVF Conference on Computer Vision and Pattern Recognition (CVPR)}, June 2021, pp. 3517--3526.

\bibitem{ATD}
L.~Zhang, Y.~Li, X.~Zhou, X.~Zhao, and S.~Gu, ``Transcending the limit of local window: Advanced super-resolution transformer with adaptive token dictionary,'' in \emph{Proceedings of the IEEE/CVF Conference on Computer Vision and Pattern Recognition}, 2024, pp. 2856--2865.

\bibitem{han2022survey}
K.~Han, Y.~Wang, H.~Chen, X.~Chen, J.~Guo, Z.~Liu, Y.~Tang, A.~Xiao, C.~Xu, Y.~Xu \emph{et~al.}, ``A survey on vision transformer,'' \emph{IEEE transactions on pattern analysis and machine intelligence}, vol.~45, no.~1, pp. 87--110, 2022.

\bibitem{Zheng_2024_CVPR}
T.~Zheng, K.~Jiang, and H.~Yao, ``Dynamic policy-driven adaptive multi-instance learning for whole slide image classification,'' in \emph{Proceedings of the IEEE/CVF Conference on Computer Vision and Pattern Recognition (CVPR)}, June 2024, pp. 8028--8037.

\bibitem{su2024high}
J.-N. Su, M.~Gan, G.-Y. Chen, W.~Guo, and C.~P. Chen, ``High-similarity-pass attention for single image super-resolution,'' \emph{IEEE Transactions on Image Processing}, vol.~33, pp. 610--624, 2024.

\bibitem{tmm10}
Q.~Liu, P.~Gao, K.~Han, N.~Liu, and W.~Xiang, ``Degradation-aware self-attention based transformer for blind image super-resolution,'' \emph{IEEE Transactions on Multimedia}, vol.~26, pp. 7516--7528, 2024.

\bibitem{Chen_2023_CVPR}
X.~Chen, H.~Li, M.~Li, and J.~Pan, ``Learning a sparse transformer network for effective image deraining,'' in \emph{Proceedings of the IEEE/CVF Conference on Computer Vision and Pattern Recognition (CVPR)}, June 2023, pp. 5896--5905.

\bibitem{hou2024linearly}
J.~Hou, Z.~Cao, N.~Zheng, X.~Li, X.~Chen, X.~Liu, X.~Cong, M.~Zhou, and D.~Hong, ``Linearly-evolved transformer for pan-sharpening,'' \emph{arXiv preprint arXiv:2404.12804}, 2024.

\bibitem{rgt}
Z.~Chen, Y.~Zhang, J.~Gu, L.~Kong, and X.~Yang, ``Recursive generalization transformer for image super-resolution,'' in \emph{ICLR}, 2024.

\bibitem{hat}
X.~Chen, X.~Wang, J.~Zhou, Y.~Qiao, and C.~Dong, ``Activating more pixels in image super-resolution transformer,'' in \emph{Proceedings of the IEEE/CVF Conference on Computer Vision and Pattern Recognition (CVPR)}, June 2023, pp. 22\,367--22\,377.

\bibitem{kalman1960new}
R.~E. Kalman, ``A new approach to linear filtering and prediction problems,'' \emph{Journal of Basic Engineering}, vol.~82, no.~1, pp. 35--45, 1960.

\bibitem{guo2022visual}
M.-H. Guo, C.-Z. Lu, Z.-N. Liu, M.-M. Cheng, and S.-M. Hu, ``Visual attention network,'' \emph{arXiv preprint arXiv:2202.09741}, 2022.

\bibitem{su2022global}
J.-N. Su, M.~Gan, G.-Y. Chen, J.-L. Yin, and C.~P. Chen, ``Global learnable attention for single image super-resolution,'' \emph{IEEE Transactions on Pattern Analysis and Machine Intelligence}, vol.~45, no.~7, pp. 8453--8465, 2022.

\bibitem{huang2023pami}
W.~Huang, M.~Ye, Z.~Shi, and B.~Du, ``Generalizable heterogeneous federated cross-correlation and instance similarity learning,'' \emph{IEEE Transactions on Pattern Analysis and Machine Intelligence}, vol.~46, no.~2, pp. 712--728, 2024.

\bibitem{FCCL_CVPR22}
W.~Huang, M.~Ye, and B.~Du, ``Learn from others and be yourself in heterogeneous federated learning,'' in \emph{Proceedings of the IEEE/CVF Conference on Computer Vision and Pattern Recognition}, 2022, pp. 10\,143--10\,153.

\bibitem{ediffsr}
Y.~Xiao, Q.~Yuan, K.~Jiang, J.~He, X.~Jin, and L.~Zhang, ``Ediffsr: An efficient diffusion probabilistic model for remote sensing image super-resolution,'' \emph{IEEE Transactions on Geoscience and Remote Sensing}, vol.~62, pp. 1--14, 2024.

\bibitem{chen2023cross}
S.~Chen, L.~Zhang, and L.~Zhang, ``Cross-scope spatial-spectral information aggregation for hyperspectral image super-resolution,'' \emph{arXiv preprint arXiv:2311.17340}, 2023.

\bibitem{srcnn}
C.~Dong, C.~C. Loy, K.~He, and X.~Tang, ``Image super-resolution using deep convolutional networks,'' \emph{IEEE transactions on pattern analysis and machine intelligence}, vol.~38, no.~2, pp. 295--307, 2015.

\bibitem{vdsr}
J.~Kim, J.~K. Lee, and K.~M. Lee, ``Accurate image super-resolution using very deep convolutional networks,'' in \emph{Proceedings of the IEEE conference on computer vision and pattern recognition}, 2016, pp. 1646--1654.

\bibitem{rdn}
Y.~Zhang, Y.~Tian, Y.~Kong, B.~Zhong, and Y.~Fu, ``Residual dense network for image super-resolution,'' in \emph{CVPR}, 2018.

\bibitem{rcan}
Y.~Zhang, K.~Li, K.~Li, L.~Wang, B.~Zhong, and Y.~Fu, ``Image super-resolution using very deep residual channel attention networks,'' in \emph{Proceedings of the European conference on computer vision (ECCV)}, 2018, pp. 286--301.

\bibitem{han}
B.~Niu, W.~Wen, W.~Ren, X.~Zhang, L.~Yang, S.~Wang, K.~Zhang, X.~Cao, and H.~Shen, ``Single image super-resolution via a holistic attention network,'' in \emph{Computer Vision--ECCV 2020: 16th European Conference, Glasgow, UK, August 23--28, 2020, Proceedings, Part XII 16}.\hskip 1em plus 0.5em minus 0.4em\relax Springer, 2020, pp. 191--207.

\bibitem{hsenet}
S.~Lei and Z.~Shi, ``Hybrid-scale self-similarity exploitation for remote sensing image super-resolution,'' \emph{IEEE Transactions on Geoscience and Remote Sensing}, vol.~60, pp. 1--10, 2022.

\bibitem{chen2023msdformer}
S.~Chen, L.~Zhang, and L.~Zhang, ``Msdformer: Multiscale deformable transformer for hyperspectral image super-resolution,'' \emph{IEEE Transactions on Geoscience and Remote Sensing}, vol.~61, pp. 1--14, 2023.

\bibitem{transenet}
S.~Lei, Z.~Shi, and W.~Mo, ``Transformer-based multistage enhancement for remote sensing image super-resolution,'' \emph{IEEE Transactions on Geoscience and Remote Sensing}, vol.~60, pp. 1--11, 2022.

\bibitem{grl}
Y.~Li, Y.~Fan, X.~Xiang, D.~Demandolx, R.~Ranjan, R.~Timofte, and L.~Van~Gool, ``Efficient and explicit modelling of image hierarchies for image restoration,'' in \emph{Proceedings of the IEEE/CVF Conference on Computer Vision and Pattern Recognition}, 2023, pp. 18\,278--18\,289.

\bibitem{mambair}
H.~Guo, J.~Li, T.~Dai, Z.~Ouyang, X.~Ren, and S.-T. Xia, ``Mambair: A simple baseline for image restoration with state-space model,'' \emph{arXiv preprint arXiv:2402.15648}, 2024.

\bibitem{smith2022simplified}
J.~T. Smith, A.~Warrington, and S.~Linderman, ``Simplified state space layers for sequence modeling,'' in \emph{International Conference on Learning Representations}, 2022.

\bibitem{gu2022efficiently}
A.~Gu, K.~Goel, and C.~Re, ``Efficiently modeling long sequences with structured state spaces,'' in \emph{International Conference on Learning Representations}, 2022.

\bibitem{mamba}
A.~Gu and T.~Dao, ``Mamba: Linear-time sequence modeling with selective state spaces,'' \emph{arXiv preprint arXiv:2312.00752}, 2023.

\bibitem{dong2024fusionmamba}
W.~Dong, H.~Zhu, S.~Lin, X.~Luo, Y.~Shen, X.~Liu, J.~Zhang, G.~Guo, and B.~Zhang, ``Fusion-mamba for cross-modality object detection,'' 2024.

\bibitem{liu2024vmamba}
Y.~Liu, Y.~Tian, Y.~Zhao, H.~Yu, L.~Xie, Y.~Wang, Q.~Ye, and Y.~Liu, ``Vmamba: Visual state space model,'' \emph{arXiv preprint arXiv:2401.10166}, 2024.

\bibitem{ma2024umamba}
J.~Ma, F.~Li, and B.~Wang, ``U-mamba: Enhancing long-range dependency for biomedical image segmentation,'' \emph{arXiv preprint arXiv:2401.04722}, 2024.

\bibitem{yang2020fda}
Y.~Yang and S.~Soatto, ``Fda: Fourier domain adaptation for semantic segmentation,'' in \emph{Proceedings of the IEEE/CVF conference on computer vision and pattern recognition}, 2020, pp. 4085--4095.

\bibitem{xu2021fourier}
Q.~Xu, R.~Zhang, Y.~Zhang, Y.~Wang, and Q.~Tian, ``A fourier-based framework for domain generalization,'' in \emph{Proceedings of the IEEE/CVF conference on computer vision and pattern recognition}, 2021, pp. 14\,383--14\,392.

\bibitem{mao2021deep}
X.~Mao, Y.~Liu, W.~Shen, Q.~Li, and Y.~Wang, ``Deep residual fourier transformation for single image deblurring,'' \emph{arXiv preprint arXiv:2111.11745}, vol.~2, no.~3, p.~5, 2021.

\bibitem{li2023embedding}
C.~Li, C.-L. Guo, Z.~Liang, S.~Zhou, R.~Feng, C.~C. Loy \emph{et~al.}, ``Embedding fourier for ultra-high-definition low-light image enhancement,'' in \emph{The Eleventh International Conference on Learning Representations}, 2023.

\bibitem{guo2023spatial}
S.~Guo, H.~Yong, X.~Zhang, J.~Ma, and L.~Zhang, ``Spatial-frequency attention for image denoising,'' \emph{arXiv preprint arXiv:2302.13598}, 2023.

\bibitem{wang2023spatial}
C.~Wang, J.~Jiang, Z.~Zhong, and X.~Liu, ``Spatial-frequency mutual learning for face super-resolution,'' in \emph{Proceedings of the IEEE/CVF Conference on Computer Vision and Pattern Recognition}, 2023, pp. 22\,356--22\,366.

\bibitem{cho2021rethinking}
S.-J. Cho, S.-W. Ji, J.-P. Hong, S.-W. Jung, and S.-J. Ko, ``Rethinking coarse-to-fine approach in single image deblurring,'' in \emph{Proceedings of the IEEE/CVF international conference on computer vision}, 2021, pp. 4641--4650.

\bibitem{aid}
G.-S. Xia, J.~Hu, F.~Hu, B.~Shi, X.~Bai, Y.~Zhong, L.~Zhang, and X.~Lu, ``Aid: A benchmark data set for performance evaluation of aerial scene classification,'' \emph{IEEE Transactions on Geoscience and Remote Sensing}, vol.~55, no.~7, pp. 3965--3981, 2017.

\bibitem{dota}
G.-S. Xia, X.~Bai, J.~Ding, Z.~Zhu, S.~Belongie, J.~Luo, M.~Datcu, M.~Pelillo, and L.~Zhang, ``Dota: A large-scale dataset for object detection in aerial images,'' in \emph{Proceedings of the IEEE conference on computer vision and pattern recognition}, 2018, pp. 3974--3983.

\bibitem{edsr}
B.~Lim, S.~Son, H.~Kim, S.~Nah, and K.~Mu~Lee, ``Enhanced deep residual networks for single image super-resolution,'' in \emph{Proceedings of the IEEE conference on computer vision and pattern recognition workshops}, 2017, pp. 136--144.

\bibitem{dior}
K.~Li, G.~Wan, G.~Cheng, L.~Meng, and J.~Han, ``Object detection in optical remote sensing images: A survey and a new benchmark,'' \emph{ISPRS journal of photogrammetry and remote sensing}, vol. 159, pp. 296--307, 2020.

\bibitem{haunet}
J.~Wang, B.~Wang, X.~Wang, Y.~Zhao, and T.~Long, ``Hybrid attention-based u-shaped network for remote sensing image super-resolution,'' \emph{IEEE Transactions on Geoscience and Remote Sensing}, vol.~61, pp. 1--15, 2023.

\bibitem{lam}
J.~Gu and C.~Dong, ``Interpreting super-resolution networks with local attribution maps,'' in \emph{Proceedings of the IEEE/CVF Conference on Computer Vision and Pattern Recognition}, 2021, pp. 9199--9208.

\bibitem{ssim}
Z.~Wang, A.~Bovik, H.~Sheikh, and E.~Simoncelli, ``Image quality assessment: from error visibility to structural similarity,'' \emph{IEEE Transactions on Image Processing}, vol.~13, no.~4, pp. 600--612, 2004.

\bibitem{lpips}
R.~Zhang, P.~Isola, A.~A. Efros, E.~Shechtman, and O.~Wang, ``The unreasonable effectiveness of deep features as a perceptual metric,'' in \emph{Proceedings of the IEEE conference on computer vision and pattern recognition}, 2018, pp. 586--595.

\bibitem{ttst}
Y.~Xiao, Q.~Yuan, K.~Jiang, J.~He, C.-W. Lin, and L.~Zhang, ``Ttst: A top-k token selective transformer for remote sensing image super-resolution,'' \emph{IEEE Transactions on Image Processing}, vol.~33, pp. 738--752, 2024.

\bibitem{vaswani2017attention}
A.~Vaswani, N.~Shazeer, N.~Parmar, J.~Uszkoreit, L.~Jones, A.~N. Gomez, {\L}.~Kaiser, and I.~Polosukhin, ``Attention is all you need,'' \emph{Advances in neural information processing systems}, vol.~30, 2017.

\end{thebibliography}


%

\newpage
\begin{IEEEbiography}[{\includegraphics[width=1in,height=1.25in,clip,keepaspectratio]{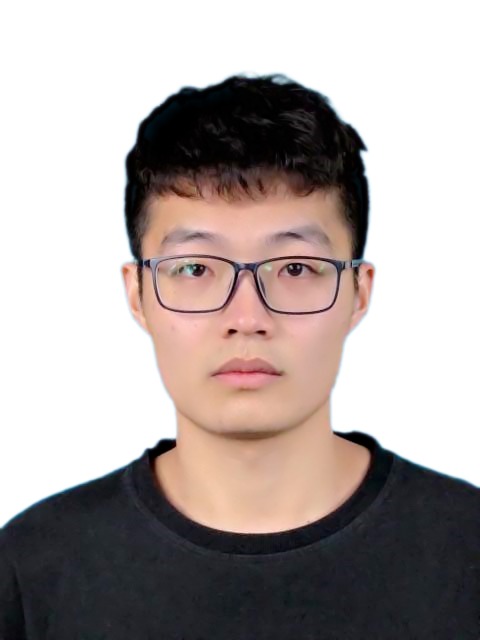}}]{Yi Xiao} (Graduate Student Member, IEEE) received the B.S. degree from the School of Mathematics and Physics, China University of Geosciences, Wuhan, China, in 2020. He is pursuing the Ph.D. degree with the School of Geodesy and Geomatics, Wuhan University, Wuhan.
\par His major research interests are remote sensing image/video processing and computer vision. More details can be found at \href{https://xy-boy.github.io/}{\textcolor{black}{https://xy-boy.github.io/}}.
\end{IEEEbiography}

\begin{IEEEbiography}[{\includegraphics[width=1in,height=1.25in,clip,keepaspectratio]{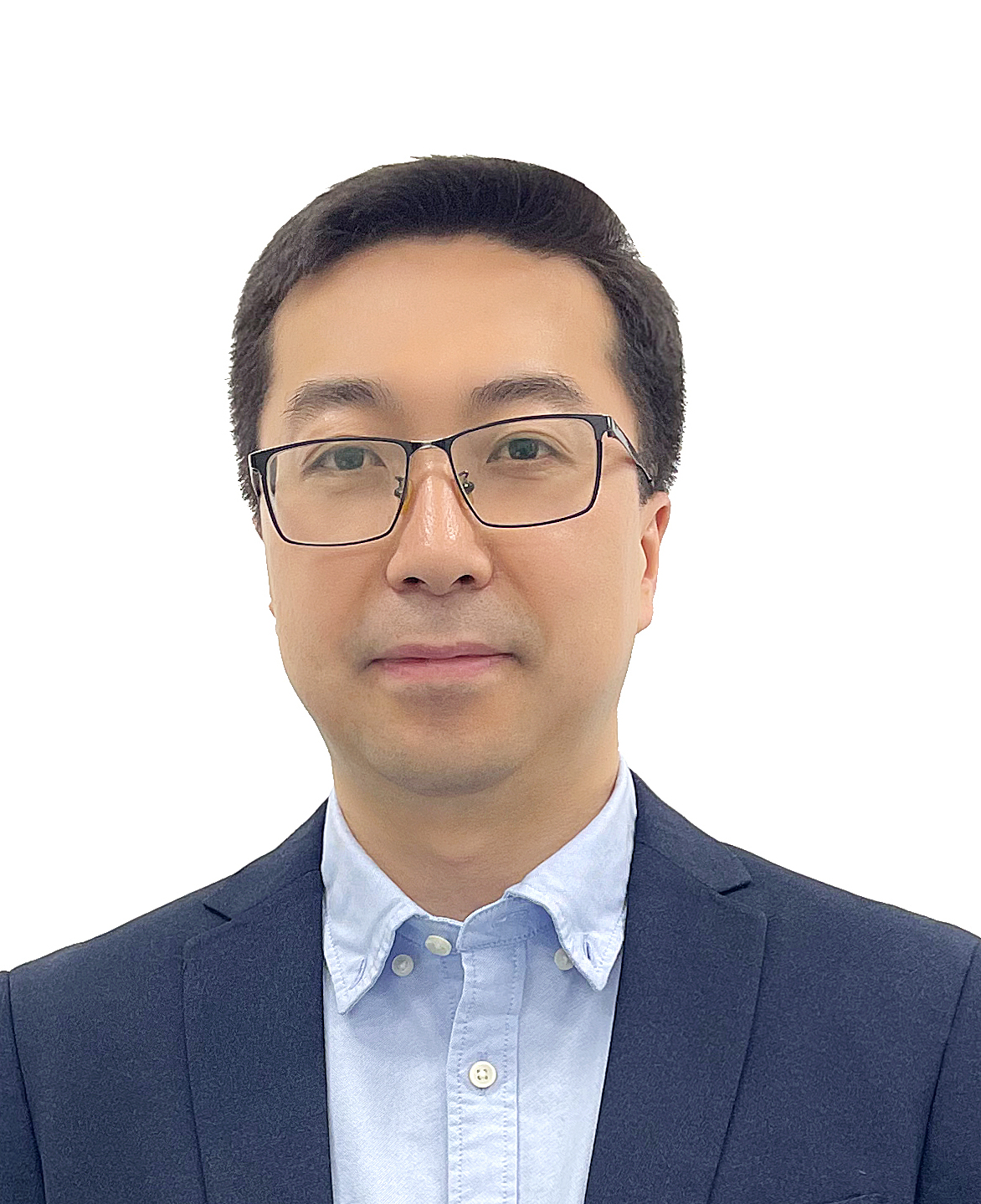}}]{Qiangqiang Yuan}
(Member, IEEE) received the B.S. degree in surveying and mapping engineering and the Ph.D. degree in photogrammetry and remote sensing from Wuhan University, Wuhan, China, in 2006 and 2012, respectively.
\par In 2012, he joined the School of Geodesy and Geomatics, Wuhan University, where he is a Professor. He has published more than 90 research papers, including more than 70 peer-reviewed articles in international journals, such as \emph{Remote Sensing of Environment, ISPRS Journal of Photogrammetry and Remote Sensing}, {\sc IEEE Transaction ON Image Processing}, and {\sc IEEE Transactions ON Geoscience AND Remote Sensing}. His research interests include image reconstruction, remote sensing image processing and application, and data fusion.
\par Dr. Yuan was a recipient of the Youth Talent Support Program of China in 2019, the Top-Ten Academic Star of Wuhan University in 2011, and the recognition of Best Reviewers of the IEEE GRSL in 2019. In 2014, he received the Hong Kong Scholar Award from the Society of Hong Kong Scholars and the China National Postdoctoral Council. He is an associate editor of 5 international journals and has frequently served as a referee for more than 40 international top journals, such as Nature Climate Change, Nature Communications, etc.
\end{IEEEbiography}

\begin{IEEEbiography}[{\includegraphics[width=1in,height=1.25in,clip,keepaspectratio]{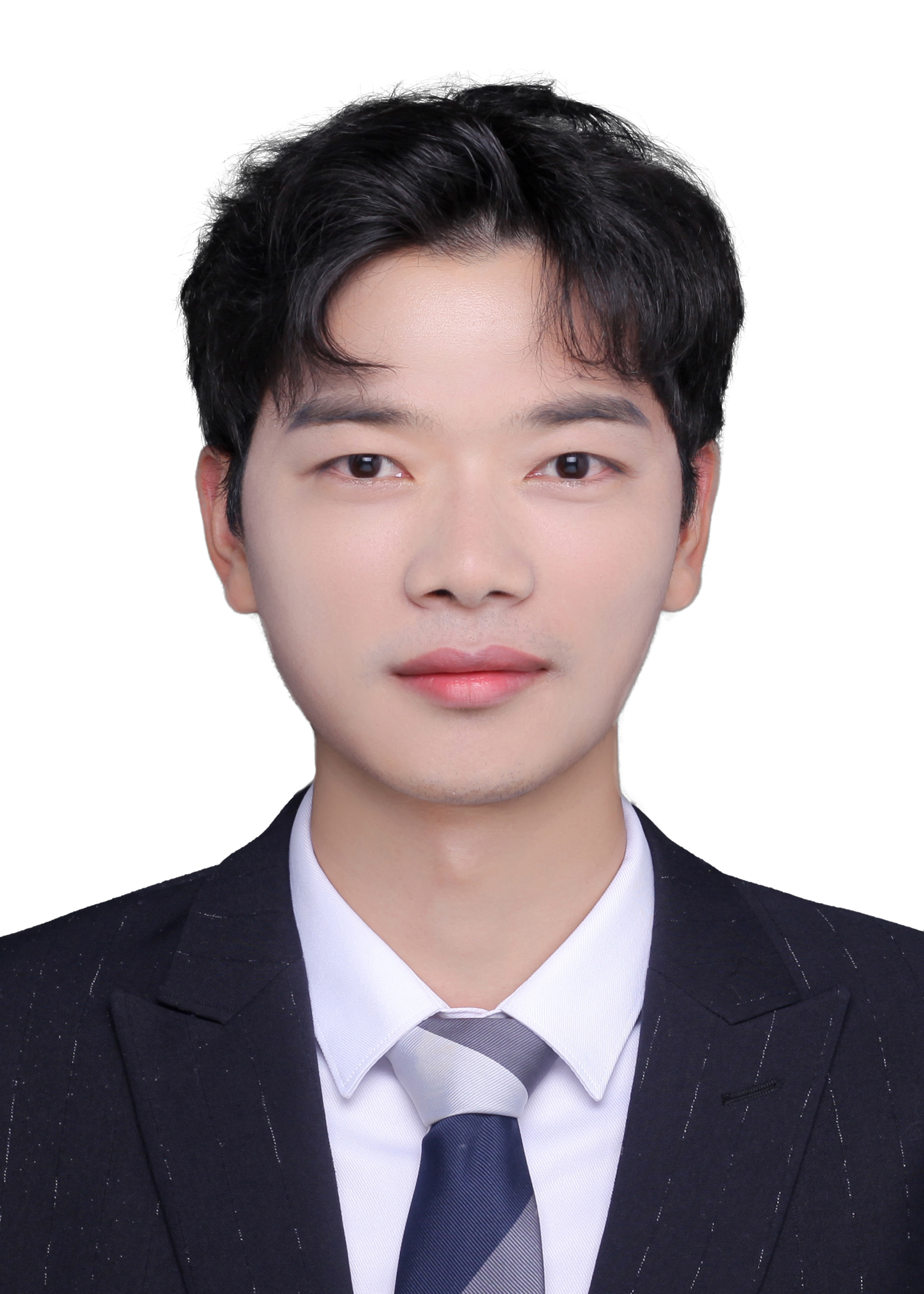}}]{Kui Jiang}
(Member, IEEE) received the M.E. and Ph.D. degrees in the School of Computer Science, Wuhan University, Wuhan, China, in 2019 and 2022, respectively. Before 2023.07, he was a research scientist with the Cloud BU, Huawei. He is currently an Associate Professor with the School of Computer Science and Technology, Harbin Institute of Technology. He received the  2022 ACM Wuhan Doctoral Dissertation Award. His research interests include image/video processing and computer vision.
\end{IEEEbiography}

\begin{IEEEbiography}[{\includegraphics[width=1in,height=1.25in,clip,keepaspectratio]{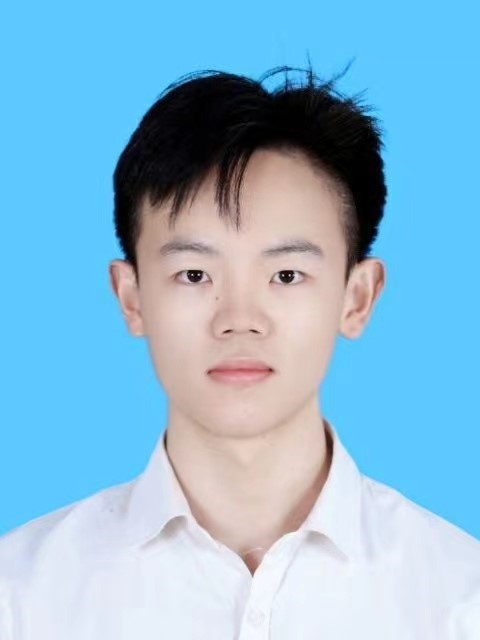}}]{Yuzeng Chen}
received the B.S. degree in geographic information science from Southwest University of Science and Technology, Mianyang, China, in 2020 and the M.S. degree in surveying and mapping engineering from Central South University in Changsha, China, in 2023. He is pursuing the Ph.D. degree with the School of Geodesy and Geomatics, Wuhan University, Wuhan. 

His research interests include remote-sensing video processing and computer vision.
\end{IEEEbiography}

\begin{IEEEbiography}[{\includegraphics[width=1in,height=1.25in,clip,keepaspectratio]{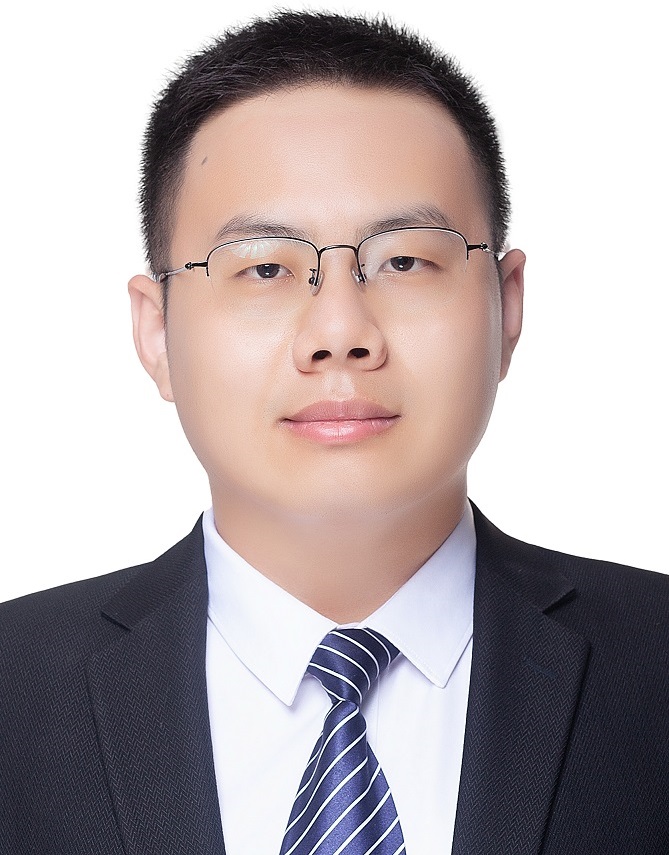}}]{Qiang Zhang} (Member, IEEE) received the B.E. degree in surveying and mapping engineering, M.E. and Ph.D. degree in photogrammetry and remote sensing from Wuhan University, Wuhan, China, in 2017, 2019 and 2022, respectively.
	\par He is currently an Associate Professor with the Center of Hyperspectral Imaging in Remote Sensing (CHIRS), Information Science and Technology College, Dalian Maritime University. His research interests include remote sensing information processing, computer vision, and machine learning. He has published more than ten journal papers on IEEE TIP, IEEE TGRS, ESSD, and ISPRS P\&RS. More details could be found at \href{https://qzhang95.github.io}{\textcolor{black}{https://qzhang95.github.io}}.
\end{IEEEbiography}

\begin{IEEEbiography}[{\includegraphics[width=1in,height=1.25in,clip,keepaspectratio]{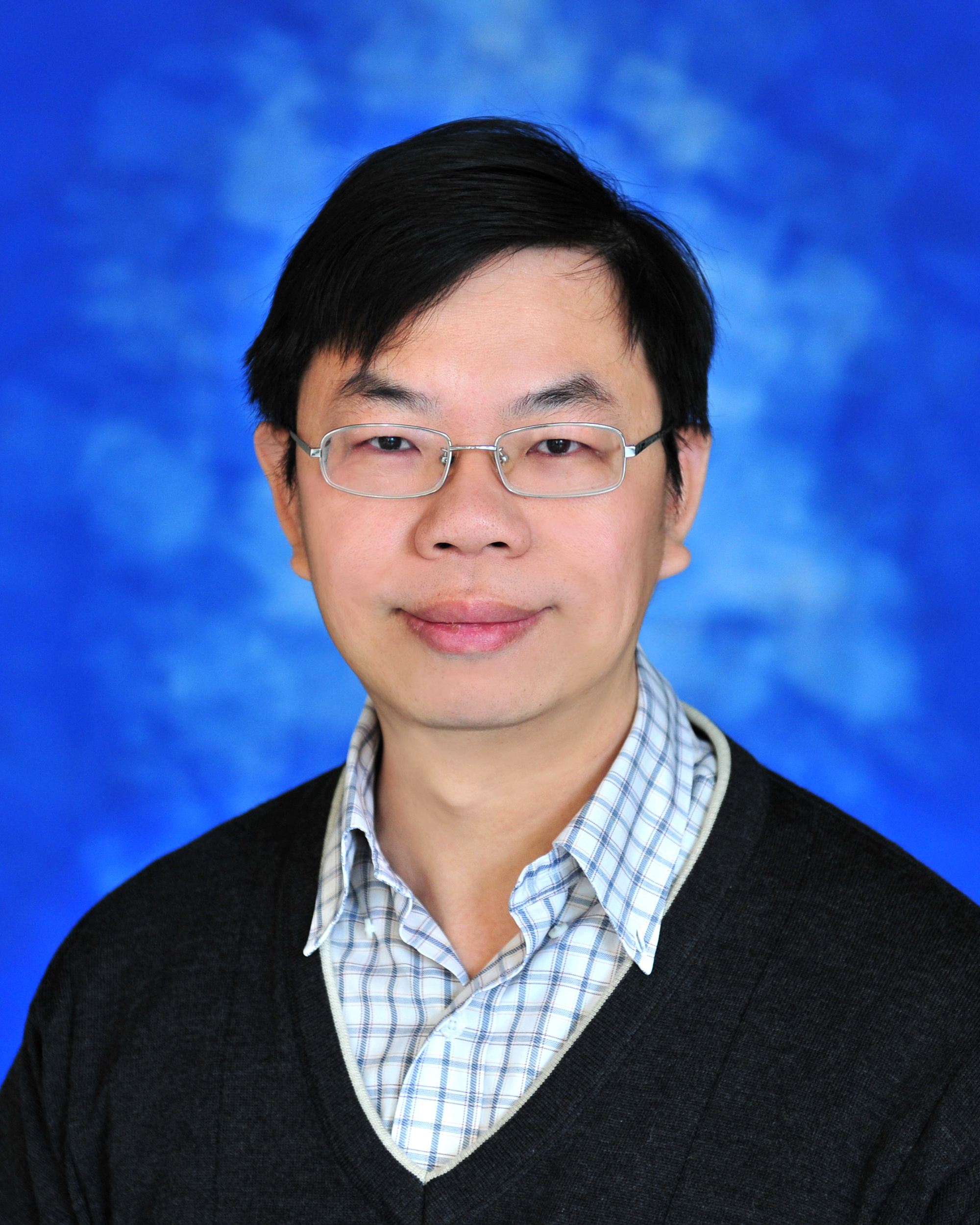}}]{Chia-Wen Lin}
(Fellow, IEEE) received the Ph.D degree in electrical engineering from National Tsing Hua University (NTHU), Hsinchu, Taiwan, in 2000. He is currently a Professor at the Department of Electrical Engineering and the Institute of Communications Engineering at NTHU. He is also Deputy Director of the AI Research Center of NTHU. He was with the Department of Computer Science and Information Engineering, National Chung Cheng University, Taiwan, during 2000–2007. Prior to joining academia, he worked for the Information and Communications Research Laboratories, Industrial Technology Research Institute, Hsinchu, Taiwan, during 1992– 2000. His research interests include image and video processing, computer vision, and video networking. 
\par Dr. Lin served as a Distinguished Lecturer of IEEE Circuits and Systems Society from 2018 to 2019, a Steering Committee member of IEEE TRANSACTIONS ON MULTIMEDIA from 2014 to 2015, and the Chair of the Multimedia Systems and Applications Technical Committee of the IEEE Circuits and Systems Society from 2013 to 2015. His articles received the Best Paper Award of IEEE VCIP 2015 and the Young Investigator Award of VCIP 2005. He received the Outstanding Electrical Professor Award presented by the Chinese Institute of Electrical Engineering in 2019, and the Young Investigator Award presented by the Ministry of Science and Technology, Taiwan, in 2006. He is also the Chair of the Steering Committee of IEEE ICME. He has served as a Technical Program Co-Chair for IEEE ICME 2010, and a General Co-Chair for IEEE VCIP 2018, and a Technical Program Co-Chair for IEEE ICIP 2019. He has served as an Associate Editor of IEEE TRANSACTIONS ON IMAGE PROCESSING, IEEE TRANSACTIONS ON CIRCUITS AND SYSTEMS FOR VIDEO TECHNOLOGY, IEEE TRANSACTIONS ON MULTIMEDIA, IEEE MULTIMEDIA, and Journal of Visual Communication and Image Representation.
\end{IEEEbiography}




\end{document}